\newcommand{\sys}{Thunder-DeID+\xspace}

\documentclass[11pt]{article}

\usepackage[final]{acl}

\usepackage{times}
\usepackage{latexsym}

\usepackage[T1]{fontenc}
\usepackage[utf8]{inputenc}

\usepackage{microtype}

\usepackage{inconsolata}

\usepackage{graphicx}

\usepackage{kotex}
\usepackage{amsmath}

\usepackage{soul} % highlight
\soulregister\sys7
\usepackage{xurl}
\usepackage{hyperref}
\usepackage{tabularx}
\usepackage{booktabs} % toprule / midrule
\usepackage{makecell}
\usepackage{xspace}
\usepackage{amssymb}
\usepackage{placeins}
\usepackage{needspace}

\usepackage{multirow} % table
\usepackage{pgf-pie} % pie chart

\usepackage{enumitem}

\title{\sys: Boundary-Aware Korean De-identification \\ from Court Judgments to Broader Sensitive Documents}

\author{
    \textbf{Gyuseong Lee}$^{1,*}$\;
    \textbf{Gyeongje Cho}$^{1,*}$\;
    \textbf{Jongmin Kim}$^{1,*}$\;
    \textbf{Heejin Kim}$^{1,*}$ \\
    \textbf{Yeonkyoung So}$^{1}$\;
    \textbf{Hyunggeun Jeon}$^{1}$\;
    \textbf{Hyunji M. Park}$^{1}$\;
    \textbf{Sangho Kim}$^{1}$ \\
    \textbf{Yoonhee Park}$^{1}$\;
    \textbf{Youngbin Pyo}$^{1}$\;
    \textbf{Jaejin Lee}$^{1,2}$ \\[2pt]
    $^{1}$Dept. of Data Science, Seoul National University \\
    $^{2}$Dept. of Computer Science and Engineering, Seoul National University \\[2pt]
    \texttt{\{ksnannaya, avq159, jmin.kim, kheejin\}@snu.ac.kr} \\
    \texttt{\{kathy1028, jhg123456, mhj233, ksh4931\}@snu.ac.kr} \\
    \texttt{\{yoonheepark, ypyo01, jaejin\}@snu.ac.kr} \\[1pt]
    {\small $^{*}$Equal contribution.
    }
}

\begin{document}
    \maketitle
    \begin{abstract}
This study examines the transferability of a de-identification model developed for Korean court judgments to a broader range of sensitive documents, focusing on boundary precision, label granularity, and recall. The agglutinative structure of Korean causes Byte Pair Encoding (BPE) tokenizers to frequently merge Personally Identifiable Information (PII) with adjacent particles, which complicates accurate span recovery. Furthermore, current PII taxonomies lack sufficient granularity for legal texts. To address these issues, this study introduces \sys, which integrates a refined three-tier PII taxonomy with a boundary-aware tokenizer. The tokenizer enhances a pretrained BPE vocabulary by incorporating N-gram and boundary-scoring components to better align token and PII span boundaries. The proposed approach is evaluated using a tokenizer-independent word-segment-level protocol. On full-length judgments, the tokenizer consistently improves W. Strict Typed F1-score from 0.76–0.77 to 0.87–0.88 across four model sizes (150M, 395M, 750M, and 1B), mainly by increasing precision rather than model scale. With fine-tuning on 1,000 broader sensitive documents, the judgment-trained model further achieves 0.93 W. Strict Typed F1-score, indicating effective transfer beyond court judgments.
\end{abstract}
    \section{Introduction}
Legal documents contain dense, context-rich information about individuals, organizations, events, disputes, transactions, and institutional decisions. In South Korea, court judgments are publicly disclosed under the legal framework governing access to them, but must be released in a manner that protects personal and case-specific identifying information. This requirement creates a central tension: court judgments should remain valuable for public access, legal research, and downstream processing, while sensitive information must be removed or masked prior to disclosure or reuse. Consequently, Korean court judgment de-identification is not merely a removal problem. The task requires reducing re-identification risk while preserving the legal content that renders judgments useful.

In Korea, where the Personal Information Protection Act covers not only direct identifiers but also information that can identify an individual when combined with other information~\citep{korea_pipa}, court-deidentified data requires more than simple PII removal. Identifiability often depends on relationships among persons, organizations, events, locations, dates, and case-specific facts, so systems must understand contextual relations rather than detect isolated entities. The task is further complicated by the Korean agglutinative structure, in which PII may be attached to particles or other morphemes without intervening whitespace, making boundary-accurate masking difficult. Since raw legal documents cannot always be sent to external LLMs before masking, reliable de-identification models must be developed for controlled or offline environments while preserving legal utility.

Previous research has addressed the de-identification of Korean court judgments using NLP methods, including Thunder-DeID and K-LegalDeID~\citep{thunder-deid, choi2026k}. However, these approaches are less suited to real-world de-identification tasks, where diverse case types and context-dependent identifiers necessitate finer-grained, more systematic PII categories, more accurate boundary modeling, and evaluation protocols that address residual privacy risks.

To address these challenges, we propose \sys, a Korean de-identification framework that extends court judgment de-identification to broader sensitive document settings. \sys integrates an annotated judgment corpus, an extensible three-tier PII taxonomy, a boundary-aware tokenization and span-alignment strategy, and a word-segment-level evaluation protocol. On full-length judgments, \sys achieves substantial improvements over Thunder-DeID in Strict Binary F1-score, with gains attributable to improved precision without loss of recall. The framework also transfers effectively to non-judgment Korean documents with minimal fine-tuning.

The main contributions of this paper are summarized as follows: 
\begin{itemize}[nosep]
    \item We redesign Korean legal PII categories into an extensible three-tier taxonomy that reduces redundant labels, resolves inconsistent granularity, and enables realistic data augmentation by replacing each PII span with a label-matched surrogate value. 
    \item We propose a tokenizer that combines an N-gram language model with a boundary-aware model to separate Korean PII spans from adjacent particles and functional morphemes, leading to consistent F1 gains over standard BPE across model sizes. 
    \item We introduce a tokenizer-independent word-segment-level evaluation protocol that assesses both boundary coverage and type correctness through Strict/Partial and Typed/Binary metric variants. 
    \item We conduct a domain transfer study from Korean court judgments to broader non-judgment sensitive documents, showing that judgment-trained \sys adapts effectively to broader Korean sensitive documents. 
\end{itemize}

% (1) PII revision -> 비식별화 모델 성능향상
% (2) tokenizer
% (3) metric (word-sgemgent-lvel evaluation)
% (4) domain transfer 
    \section{Related Work}

\paragraph{De-identification in diverse domains.}
Automatic de-identification systems such as Scrub and Datafly have been primarily developed in the medical domain~\citep{sweeney1996replacing,sweeney1998datafly}, where sensitive records are shared under privacy regulations such as HIPAA and the GDPR~\citep{hhs1996hipaa,meystre2010automatic,gdpr2016,lison2021anonymisation}. 
Subsequent research has adopted BiLSTM-CRF models, transformer encoders, and, more recently, LLM-based taggers or teacher models in this task~\citep{dernoncourt2017identification,liu2017identification,yang2019study,tang2020identification,johnson2020deidentification,vakili2024end,liu2023deid,yashwanth2024zero,altalla2025evaluating,deusser2025resource}.
Beyond medicine~\citep{uzuner2007evaluating,stubbs2015annotating}, automatic de-identification has also been studied in law and court rulings~\citep{glaser2021anonymization,oksanen2022anonymization,pilan2022text,niklaus2023automatic,itani2024automated}, finance~\citep{biesner2022anonymization,mishra2025hybrid}, emails~\citep{eder2019identification}, recruitment~\citep{van2020dutch,jensen2021identification}, and other specialized contexts~\citep{shahriar2024prism,abdjul2025indonesian,papadopoulou2026neural}. 

% In these settings, the scope of identifying information is influenced by legal requirements, document structure, and domain-specific re-identification risks. This challenge is particularly pronounced in court judgments, where workplaces, institutions, incident locations, and other contextual attributes may collectively identify parties even after direct identifiers such as names and registration numbers have been removed.

\paragraph{De-identification of Korean court judgments.}
For Korean court judgments, two recent studies have treated de-identification as a supervised learning problem: Thunder-DeID~\citep{thunder-deid} and K-LegalDeID~\citep{choi2026k}. 

Thunder-DeID introduced the first annotated Korean court judgment corpus, comprising 6,700 civil, criminal, and administrative judgments with 48,306 annotated entities, and proposed a three-tier taxonomy of personally identifiable information (PII) with 729 fine-grained labels. The system combined Mecab-ko\footnote{https://github.com/hephaex/mecab-ko}, a Korean morphological analyzer, with BPE~\citep{sennrich2015neural}, a subword segmentation method, to more effectively separate entity spans from adjacent Korean particles. While this design addresses a significant Korean boundary problem, it also introduces a trade-off. A highly fine-grained label space results in many sparse prediction targets, and morpheme-aware tokenization can increase sequence length, thereby raising inference costs and reducing compatibility with pretrained Korean encoders.

% K-LegalDeID employs a more compact design. It covers 39 legal fields, uses BIO tagging with a conditional random field (CRF) layer on top of KLUEBERT~\citep{park2021klue}, and consolidates PII into eleven flat categories. This approach simplifies the prediction task but also merges many indirect identifiers, or quasi-identifiers. These are attributes that may not, on their own, identify a party, but may increase re-identification risk when combined with other facts in the judgment. In court judgments, typical examples include workplaces, institutional affiliations, and incident locations.

K-LegalDeID builds on the Thunder-DeID dataset~\citep{thunder-deid} but adopts a substantially simplified labeling scheme. It covers 39 legal fields, uses BIO tagging with a conditional random field (CRF) layer on top of KLUEBERT~\citep{park2021klue}, and consolidates the PII annotations into eleven flat categories. While this reduces the complexity of the prediction task, it comes at the cost of collapsing many indirect identifiers, or quasi-identifiers, distinguished in Thunder-DeID. Such attributes may not identify a party on their own but can increase re-identification risk when combined with other facts in the judgment, including workplaces, institutional affiliations, and incident locations.

% The proposed approach maintains a hierarchical taxonomy for judgment-specific indirect identifiers, but redesigns the label space so that related categories are defined at comparable levels of abstraction and adhere to a single classification criterion. A tokenizer is also introduced that preserves Korean entity-boundary alignment while reducing sequence length and improving compatibility with pretrained Korean encoders. Both in-domain judgment performance and generalization to non-judgment Korean documents are evaluated.

Building on prior work on Korean legal de-identification, this work presents a de-identification system with a redesigned hierarchical taxonomy that covers indirect identifiers and a length-efficient tokenizer that preserves entity-boundary alignment in Korean. It also examines generalization beyond court judgments.

    \section{Methods}

\subsection{Court Judgment Corpus Construction} % KS 수정 (Dataset -> Court judgment Corpus Cosntruction)
\label{sec:Method:03_sub_corpus_construction}
% Our corpus construction procedure largely follows the prior work~\cite{thunder-deid}, with minor modifications.

\paragraph{Data collection.} 
We collected 27,992 court judgments from datasets provided by the Korean Ministry of Government Legislation\footnote{https://www.moleg.go.kr/} and the Judicial Information Disclosure Portal of the Courts of the Republic of Korea\footnote{https://portal.scourt.go.kr/}. In contrast to prior work~\cite{thunder-deid}, which primarily used main-text excerpts instead of complete court judgments, we prioritized collecting full judgments whenever available. This approach enables our dataset to retain richer document-level context.

We follow the overall construction workflow established in prior work~\cite{thunder-deid}, but introduce a newly designed PII taxonomy (Section~\ref{sec:Method:03_sub_PIItaxonomy}). The construction process includes OCR and boilerplate removal for scanned judgments, full-document annotation with contextual de-identification guidelines, and line-break normalization without changing the document content. Additional implementation details are available in Appendices~\ref{app:ocr}–\ref{app:normalization}.

\paragraph{Corpus statistics.}
Following annotation and quality filtering, the final judgment corpus contains 5,872 documents and 193,004 annotated spans. 
% \hl{The final corpus was constructed by selecting approximately 6,000 documents for annotation from the initial pool of 27,992 judgments, followed by annotation and quality filtering. This process resulted in 5,872 documents with 193,004 annotated spans.}
Relative to the Thunder-DeID corpus~\citep{thunder-deid}, our corpus encompasses a broader range of settings (Table~\ref{tab:v1-v2-dataset-difficulty}). It includes 2.8 times more fine-grained case types, longer documents, and a greater number of PII mentions per document. These characteristics increase task complexity, as systems must detect and classify more spans in longer documents across a wider array of legal sub-areas. Detailed document- and span-level statistics, as well as the full case-type distribution, are provided in Appendix~\ref{app:corpus_stats}.

\begin{table}[t]
\centering
\small
\setlength{\tabcolsep}{6pt}
\resizebox{\columnwidth}{!}{%
\begin{tabular}{lrrr}
\toprule
& Thunder-DeID & \sys & Ratio \\
\midrule
Documents & 6,700 & 5,872 & 0.88$\times$ \\
Case types & 10 & 28 & 2.80$\times$ \\
Avg.\ body chars / doc & 2,627 & 4,554 & 1.73$\times$ \\
Avg.\ PII mentions / doc & 21.0 & 32.9 & 1.56$\times$ \\
\bottomrule
\end{tabular}%
}
\caption{Comparison between the existing Thunder-DeID judgment corpus and the present corpus. All values are computed from the raw annotated judgments before replacement and split assignment. Document length is measured after removing inline annotation markup. The present corpus offers broader case coverage, longer documents, and more PII mentions per document.}
\label{tab:v1-v2-dataset-difficulty}
\vspace{-8pt}
\end{table}

\subsection{PII Taxonomy}
\label{sec:Method:03_sub_PIItaxonomy}

\begin{figure*}[ht]
    \centering
    \includegraphics[width=0.9\linewidth]{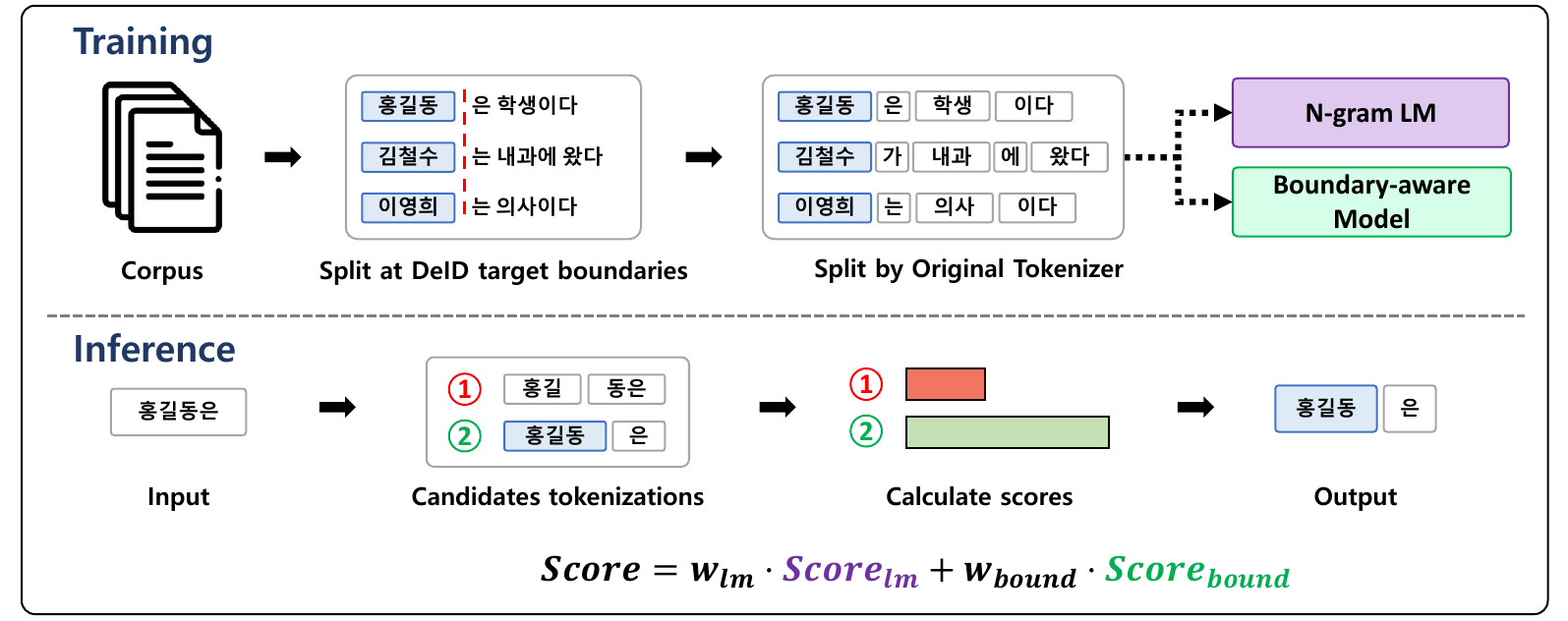}
    \caption{Illustration of the tokenizer training and inference process for accurately separating de-identification targets. Blue blocks indicate de-identification targets.}
    \label{fig:tokenizer_procedure}
\vspace{-6pt}
\end{figure*}

A central component of the de-identification pipeline is a PII taxonomy that supports both reliable annotation and realistic augmentation. Since augmentation replaces anonymized spans with plausible substitute entities, the taxonomy must be specific enough to choose replacements that fit the original context, while avoiding overly narrow PII labels that fragment similar entities and make learning harder.

The taxonomy employed in recent work~\cite{thunder-deid} exhibited several limitations in this respect.
First, many labels were excessively fragmented or partially redundant. For example, labels such as "식당", "음식점", and "레스토랑" were treated as separate PII labels, although they all denote the same broad restaurant category in Korean. 
Second, the labels varied in granularity; some entities were annotated at a broad category level, whereas others were assigned to highly specific subtypes.
This inconsistency introduced ambiguity in target labels for model training and evaluation, as the required level of specificity for correct predictions was not uniform across examples.
Third, the taxonomy conflated multiple classification criteria within single categories. For example, in the business domain, industry type (such as "eating and drinking places" or "wholesale and retail trade") and legal or organizational form (such as "corporation" or "limited liability company") were treated as competing labels, despite representing orthogonal properties of the same entity.
As a result, annotators were required to select among labels that did not follow a single, consistent classification criterion. If an entity is assigned to an unsuitable label, augmentation may sample a replacement entity that does not fit the surrounding context. 

To resolve these challenges, we redesign the PII taxonomy as a three-tier hierarchical annotation schema. The complete taxonomy is provided in Appendix~\ref{app:PIItaxonomy}. In the refined taxonomy, each broad PII category is divided into subcategories at the same level of detail, and subcategories under the same parent category are designed to cover the relevant entity types without overlapping in meaning. This makes the required level of specificity more consistent across labels. The taxonomy also assigns labels based on the role of an entity in the case context, rather than only on its surface form. Therefore, the same surface expression may receive different labels in different documents when it plays different roles in the case narrative.  
This design reduces label overlap, makes annotation decisions easier to interpret, and allows new entity types to be added without reorganizing the entire taxonomy. It further supports realistic augmentation with plausible substitute entities and provides a reusable label space for de-identification beyond court judgments.

% 데이터셋 비율 ~~ 맞췄는데 + why (사법연감) + overfittign 방지 얘기

% % KS 추가
\subsection{Tokenizer}
We propose a tokenization method for Korean de-identification that preserves the pretrained BPE vocabulary while separating PII spans from adjacent particles and endings. This design is motivated by the agglutinative structure of Korean, where named entities are frequently attached to functional morphemes without explicit whitespace. For example, in "홍길동은", the PII span is the name "홍길동", while "은" is a grammatical particle attached to it. A tokenizer may split the expression into "홍길" and "동은", where "동은" incorrectly merges the final syllable of the name with the particle. In such cases, token-level de-identification cannot precisely mask only the target entity, potentially resulting in either over-masking of the surrounding particle or under-masking of only part of the entity.

One straightforward solution to this problem is to use character-level or morpheme-level tokenization~\cite{park2021klue, an2024exaone, thunder-deid}. However, this approach increases sequence length and inference cost and may diminish the effectiveness of pretrained models whose vocabularies were constructed with the original tokenizer. 

We instead construct a tokenizer with two components, an N-gram language model and a boundary-aware model. The N-gram language model encourages token sequences similar to those produced by the original tokenizer, while the boundary-aware model encourages token boundaries to match PII span boundaries. This allows the tokenizer to preserve the original BPE vocabulary in general contexts while adjusting boundaries near PII spans. 

We train these components using only the judgment training set. The corpus is first split at PII span boundaries, and each resulting segment is tokenized using the original tokenizer. The resulting token sequences are used to train an N-gram language model with KenLM~\cite{heafield2011kenlm}. In parallel, we collect statistics about token patterns that frequently appear near PII span boundaries and use them to construct the boundary-aware model. 

At inference time, the tokenizer generates multiple candidate tokenization sequences for a given input using the original vocabulary. Each candidate is evaluated using a combined score that incorporates both the N-gram language model and the boundary-aware model. The N-gram language model favors candidates consistent with the original tokenizer's tokenization behavior, while the boundary-aware model promotes candidates whose token boundaries more closely align with the PII span boundaries. The tokenizer selects the candidate with the highest combined score as the final tokenization. This approach preserves the original tokenizer's behavior in general contexts, while selectively adjusting token boundaries around named entities when doing so enhances de-identification. A detailed formulation of the scoring function is provided in Appendix~\ref{app:tokenization_scoring}.

Figure~\ref{fig:tokenizer_procedure} presents the procedure using a Korean de-identification example. During training, sentences containing de-identification targets such as "홍길동", "김철수", and "이영희" are split at the target boundaries, after which the original tokenizer processes each resulting segment. These token sequences are then used to train the N-gram language model, while boundary-related token patterns are collected for the boundary-aware model. During inference, when provided with an input such as "홍길동은", the tokenizer evaluates alternative segmentations, such as "홍길 / 동은" and "홍길동 / 은". The former segmentation may align more closely with the original tokenizer, whereas the latter better preserves the boundary between the name and the particle. The final tokenization is determined by combining the N-gram language model score with the boundary-aware model score, enabling the tokenizer to remain consistent with the original pretrained vocabulary while producing boundaries that are more appropriate for de-identification.

% This approach allows the tokenizer to remain close to the original pretrained tokenizer in general text, while modifying tokenization behavior around de-identification targets. As a result, named entities can be masked more precisely without fully retraining the pretrained model or switching to an inefficient fine-grained tokenizer.

% 왜 중요/필요한지 강조

% training method 
% overfit 방지 -> count 기반으로. word(spcaing) 기준 n gram 2-gram / 3-gram
% 바로 앞 / 바로 뒤에 A 단어가 나올 확률이 나오게 한다음  threshold 는 휴리스틱으로 정한다고 함.
%,,,,,,,,,,,,,,,,,,,,,,,,,,,,,,,,,,,,

% replacement list josn + geneartion 둘다 썼다. 
% data augmentation 어떻게 했는지0
% \subsection{Replacement-based Training Data Generation}
% \input{sec/Method/03_sub_training_data}

% \subsection{General Document Dataset for Domain Shift}
% \subsection{Cross-domain Corpus}
% \label{sec:Method:03_sub_general_docs}
% \input{sec/Method/03_sub_general_docs}

\subsection{Word-segment-level Evaluation}
\label{sec:metric}

De-identification and NER are commonly evaluated using character-level or token-level spans. However, both approaches present challenges when comparing systems with different tokenizers. Character-span scoring assigns more weight to longer mentions even when each mention corresponds to a single PII item. For example, masking the 10-character "한국정보통신기술협회" contributes more to the score than masking the 4-character "삼성전자" solely because it contains more characters. 
Token-span scoring introduces variability based on how each tokenizer segments the same text, as shown by the "홍길동은" example, where token boundaries may or may not align with the name-particle boundary.

To avoid these issues, we evaluate predictions on tokenizer-independent word-segments. We define a \textit{word-segment} as either a gold PII span, which takes precedence even if it contains whitespace, or a whitespace-delimited unit otherwise. For example, in "홍길동은 삼성 전자 직원이다" ("Hong Gil-dong is an employee of Samsung Electronics"), the evaluation word-segments are "홍길동" (a person, PII), "은" (a topic particle, non-PII), "삼성 전자" (a company, PII; kept as one word-segment despite the internal space), and "직원이다" ("is an employee", non-PII). The word-segment more accurately reflects the content affected by masking and avoids dependence on tokenizer granularity.

We report four evaluation variants by combining two coverage criteria with two label criteria. For coverage, \emph{Strict} requires the prediction to cover the full non-space content of a gold PII word-segment, whereas \emph{Partial} requires only any non-space overlap. For label matching, \emph{Typed} requires the predicted PII type to match the gold type, whereas \emph{Binary} considers only whether a segment is PII or non-PII and ignores the type. This yields four variants: Strict Typed, Partial Typed, Strict Binary, and Partial Binary. 
For each variant, precision, recall, and F1-score are computed from true positives (TP), false positives (FP), and false negatives (FN) at the word-segment level. Multiple-seed results are micro-aggregated by summing TP, FP, and FN before recomputing the final scores. Further details are provided in Appendix~\ref{app:metric}.
% Full metric definitions and edge cases are provided in Appendix~\ref{app:metric}.

We use Word-segment-level Strict Typed F1-score (W. Strict Typed F1-score) as the primary metric because it requires both boundary-complete PII recovery and correct type classification. We also report W. Strict Binary, W. Partial Typed, and W. Partial Binary F1-scores as complementary metrics: Binary variants measure PII detection regardless of entity type, while Partial variants relax the boundary requirement to any non-space overlap.

    \section{Experiments}

\subsection{Experimental Setup}
\label{sec:exp_setup}

\begin{table*}[h]
\centering
\begingroup
\small
\setlength{\tabcolsep}{4pt}
\renewcommand{\arraystretch}{1.12}
\resizebox{\textwidth}{!}{%
\begin{tabular}{l l rrr rrr rrr rrr}
\toprule
\multirow{2}{*}{\textbf{Model}} & \multirow{2}{*}{\textbf{Tokenizer}}
& \multicolumn{3}{c}{\textbf{W. Strict Typed}}
& \multicolumn{3}{c}{\textbf{W. Partial Typed}}
& \multicolumn{3}{c}{\textbf{W. Strict Binary}}
& \multicolumn{3}{c}{\textbf{W. Partial Binary}} \\
\cmidrule(lr){3-5}\cmidrule(lr){6-8}\cmidrule(lr){9-11}\cmidrule(lr){12-14}
& & \textbf{F1} & \textbf{P} & \textbf{R} & \textbf{F1} & \textbf{P} & \textbf{R} & \textbf{F1} & \textbf{P} & \textbf{R} & \textbf{F1} & \textbf{P} & \textbf{R} \\ \midrule
% \midrule
% T-370M & Mecab-ko+BPE
% & n/a & n/a & n/a & n/a & n/a & n/a
% & 0.613 & 0.545 & 0.700 & 0.688 & 0.588 & 0.831 \\
% T-800M & Mecab-ko+BPE
% & n/a & n/a & n/a & n/a & n/a & n/a
% & 0.697 & 0.601 & 0.831 & 0.738 & 0.621 & 0.908 \\
% T-1.5B & Mecab-ko+BPE
% & n/a & n/a & n/a & n/a & n/a & n/a
% & 0.720 & 0.630 & 0.839 & 0.761 & 0.650 & 0.916 \\
% \addlinespace[2pt]\midrule
150M & BPE
& 0.760 & 0.661 & 0.894 & 0.779 & 0.670 & 0.932
& 0.785 & 0.685 & 0.920 & 0.802 & 0.692 & 0.953 \\
150M & \sys tokenizer
& 0.867 & 0.851 & 0.884 & 0.886 & 0.856 & 0.919
& 0.890 & 0.877 & 0.904 & 0.908 & 0.881 & 0.937 \\
\addlinespace[2pt]
395M & BPE
& 0.761 & 0.669 & 0.883 & 0.781 & 0.678 & 0.920
& 0.783 & 0.689 & 0.905 & 0.799 & 0.697 & 0.937 \\
395M & \sys tokenizer
& 0.876 & \textbf{0.870} & 0.881 & 0.897 & \textbf{0.875} & 0.921
& 0.897 & \textbf{0.894} & 0.899 & 0.917 & \textbf{0.898} & 0.936 \\
\addlinespace[2pt]
750M & BPE
& 0.770 & 0.675 & 0.895 & 0.790 & 0.684 & 0.935
& 0.792 & 0.696 & 0.918 & 0.809 & 0.704 & 0.952 \\
750M & \sys tokenizer
& 0.877 & 0.864 & 0.891 & \textbf{0.900} & 0.870 & 0.932
& 0.901 & 0.892 & 0.911 & \textbf{0.922} & 0.896 & 0.949 \\
\addlinespace[2pt]
1B & BPE
& 0.771 & 0.671 & \textbf{0.906} & 0.790 & 0.680 & \textbf{0.942}
& 0.791 & 0.689 & \textbf{0.927} & 0.806 & 0.696 & \textbf{0.957} \\
1B & \sys tokenizer
& \textbf{0.878} & 0.856 & 0.901 & 0.898 & 0.861 & 0.938
& \textbf{0.902} & 0.882 & 0.923 & 0.918 & 0.885 & 0.954 \\
\bottomrule
\end{tabular}%
}
\endgroup
\caption{Results on the Korean court judgment test set. Scores are averaged over three runs with different random seeds. P stands for precision and R stands for recall. Strict requires full non-space coverage of a gold PII word-segment; Partial requires any non-space coverage. Typed additionally requires the predicted PII type to match the gold type; Binary ignores type. Best scores in each metric column are shown in bold.}
\label{tab:main-indomain-results}
\vspace{-6pt}
\end{table*}

We formulate de-identification as BIO sequence labeling under the refined PII taxonomy. Character-level PII spans are converted into token-level labels using each tokenizer's offset mapping. In the BIO format, a token overlapping a PII span is labeled as \texttt{B-} if it contains the span's start offset and as \texttt{I-} otherwise, while tokens outside any PII span are labeled as \texttt{O}.

We employ ModernBERT-based encoder models~\cite{modernbert}, \sys, in four configurations (\textsc{150M}, \textsc{395M}, \textsc{750M}, \textsc{1B}). We adopt an encoder-only architecture so that the model tags only PII spans while leaving the remaining judgment text unchanged. All models are pre-trained from scratch on a balanced Korean-English corpus. Details of the architecture sizes, token budgets, and optimization settings are provided in Appendix~\ref{app:training}.

Additionally, we report results for Thunder-DeID~\cite{thunder-deid}, a previous system for Korean court judgment de-identification, as a reference. We do not include K-LegalDeID~\citep{choi2026k} in the empirical comparison because at the time of our experiments, no public checkpoint was available for direct evaluation. We evaluate the publicly released checkpoints without modification. Due to differences in label inventories, we report only binary entity-agnostic metrics for both the judgment test set and the general sensitive document benchmark.

The judgment corpus described in Section~\ref{sec:Method:03_sub_corpus_construction} is partitioned at the document level into 8:1:1 train, validation, and test splits with case-type stratification, yielding 4,698, 587, and 587 documents, respectively. 
For training and evaluation, each annotated PII span is replaced with a randomly sampled type-compatible surrogate value that matches its PII type. The detailed procedure is described in Appendix~\ref{app:replacement_augmentation}.

\subsection{Experimental Results}

\paragraph{Court judgments.}

We first evaluate \sys on the constructed judgment de-identification test set. The results are summarized in Table~\ref{tab:main-indomain-results}.
On full-length judgments, the 1B model with the \sys tokenizer achieves 0.954 W. Partial Binary recall and 0.901 W. Strict Typed recall, indicating that it successfully recovers most PII word-segments in the documents. Precision reflects how well the model de-identifies only the spans that need to be masked, thereby preserving the information in the original document. 
%Precision also remains high, reaching 88.5\% under partial matching and 85.6\% when both boundaries and labels must be correct. 
The same model reaches 0.885 W. Partial Binary precision and 0.856 W. Strict Typed precision.
Considering both precision and recall, the model achieves 0.918 W. Partial Binary F1-score for detecting at least part of a PII span and 0.878 W. Strict Typed F1-score under complete-boundary and type-correct evaluation. Although W. Strict Typed precision is highest for the 395M model, recall generally improves with model size, resulting in the highest W. Strict Typed F1-score for the 1B model.

\begin{table}[t]
\centering
\footnotesize
\setlength{\tabcolsep}{3.6pt}
\resizebox{\columnwidth}{!}{%
\begin{tabular}{lrrrrrr}%{@{\extracolsep{\fill}}lrrrrrr@{}}
\toprule
\multirow{2}{*}{\textbf{Model}}
& \multicolumn{3}{c}{\textbf{W. Strict Binary}} 
& \multicolumn{3}{c}{\textbf{W. Partial Binary}} \\
\cmidrule(lr){2-4}\cmidrule(lr){5-7}
& \textbf{F1} & \textbf{P} & \textbf{R} & \textbf{F1} & \textbf{P} & \textbf{R} \\
\midrule
Thunder-DeID(370M)
& 0.600 & 0.539 & 0.676 & 0.677 & 0.583 & 0.807 \\
Thunder-DeID(800M)
& 0.698 & 0.605 & 0.825 & 0.737 & 0.625 & 0.897 \\
Thunder-DeID(1.5B)
& 0.733 & 0.657 & 0.829 & 0.774 & 0.676 & 0.904 \\
\midrule
Thunder-DeID+(150M) 
& 0.874 & 0.849 & 0.901 & 0.895 & 0.854 & 0.939 \\
Thunder-DeID+(395M)
& 0.888 & 0.875 & 0.902 & 0.909 & 0.879 & 0.941 \\
Thunder-DeID+(750M)
& 0.896 & \textbf{0.882} & 0.909 & \textbf{0.918} & \textbf{0.887} & 0.951 \\
Thunder-DeID+(1B)
& \textbf{0.898} & 0.870 & \textbf{0.928} & 0.913 & 0.873 & \textbf{0.957} \\
\bottomrule
\end{tabular}
}
\caption{Performance on the subset of the judgment test set corresponding to the ten case types used by Thunder-DeID: Indecent Act by Compulsion, Delivery of Building, Payment, Road Traffic Act Violations, Fraud, Bodily Injury, Compensation for Damages, Lease, Assault, and Administrative cases. The subset contains 288 documents. Thunder-DeID uses the Mecab-ko+BPE tokenizer, whereas \sys uses the \sys tokenizer. Best scores in each metric column are shown in bold.}
\label{tab:thunder-deid-data-exp}
\vspace{-8pt}
\end{table}

We further compare \sys with Thunder-DeID, the strongest publicly available Korean de-identification model. Because Thunder-DeID uses a different PII taxonomy and was trained on different judgment types, we compare the models only on the judgment types used for Thunder-DeID training and report W. Strict Binary and W. Partial Binary precision, recall, and F1-score. The results are shown in Table~\ref{tab:thunder-deid-data-exp}. Compared with Thunder-DeID, \sys improves W. Partial Binary recall by approximately 5 percentage points and W. Strict Binary recall by about 10 percentage points. Precision improves even more substantially, with gains of roughly 20 percentage points in both settings. As a result, while Thunder-DeID achieves 0.733 W. Strict Binary F1-score, \sys reaches 0.898, showing a large performance gap. These results suggest that \sys achieves the strongest performance among currently available Korean court-judgment de-identification systems.

We also include four instruction-tuned open-weight LLMs as additional baselines:
%that can be deployed without sending raw judgment documents to external API services: 
Qwen 3.5 2B~\cite{qwen3.5}, Gemma 4 E2B and Gemma 4 26B-A4B~\cite{gemmateam2026gemma4technicalreport}, and Qwen 3.6 27B~\cite{qwen3.6-27b}. We evaluate all four models on the same 587-document judgment test set. Because generative LLMs may rewrite or omit parts of the input, which can alter information in the original judgment~\cite{thunder-deid}, we report source preservation together with W. Strict Typed F1. Source preservation is the percentage of documents whose generated text exactly matches the input after removing valid predicted PII tags. The results are shown in Table~\ref{tab:open-weight-llm}.

\begin{table}[t]
\centering
\footnotesize
\renewcommand{\arraystretch}{1.05}
\begin{tabular*}{\columnwidth}{@{\extracolsep{\fill}}lrr@{}}
\toprule
\textbf{Model}
& \shortstack{\textbf{Source}\\\textbf{preservation (\%)}}
& \shortstack{\textbf{W. Strict}\\\textbf{Typed F1}} \\
\midrule
Qwen 3.5 2B
& 0.00 & 0.004 \\
Gemma 4 E2B
& 25.89 & 0.141 \\
Gemma 4 26B-A4B
& 27.77 & 0.184 \\
Qwen 3.6 27B
& \textbf{81.09} & \textbf{0.483} \\
\bottomrule
\end{tabular*}
\caption{Performance of instruction-tuned open-weight LLMs on the 587-document judgment test set, reported using W. Strict Typed F1 and source preservation.}
\label{tab:open-weight-llm}
\end{table}

\begin{table}[tbp]
\centering
\footnotesize
\setlength{\tabcolsep}{7pt}
\begin{tabular*}{\columnwidth}{@{\extracolsep{\fill}}lrrr@{}}
\toprule
\textbf{Tokenizer} 
& \textbf{Vocab.}
& \textbf{Tok./1K$\downarrow$}
& \textbf{Clean cov.$\uparrow$} \\
\midrule
Unicode Char    & 2.5K  & 1000.0 & 100.0 \\
BPE    & 50K  & \textbf{459.5} & 70.8 \\
Mecab-ko+BPE  & 32K  & 539.1 & 92.4 \\
Mecab-ko+BPE & 128K & 509.0 & 90.7 \\
\sys   & 50K  & 480.2 & \textbf{99.7} \\      % vocab size 50,368임!!
\bottomrule
\end{tabular*}
\caption{Tokenizer-only boundary analysis on the in-domain judgment test split. Tok./1K is the number of tokens per 1,000 input characters, and Clean cov. is the percentage of gold PII spans covered without extra non-space characters. Vocabulary sizes are rounded. Best values are shown in bold.}
\label{tab:tokenizer-boundary-analysis}
\vspace{-8pt}
\end{table}

Qwen 3.6 27B~\cite{qwen3.6-27b} achieves the highest performance among the evaluated open-weight LLMs, but its 0.483 W. Strict Typed F1 remains substantially below the 0.867 achieved by the smallest 150M \sys model, despite its much larger model size. It also fails to preserve the source text exactly in 18.91\% of the judgments. These results show that the evaluated open-weight LLMs are substantially less effective than \sys for full-document de-identification. Details of the prompting, chunking, and source-preservation validation are provided in Appendix~\ref{app:open-weight-llm}.
To further examine how performance varies across individual PII types, we provide a per-type breakdown and error analysis in Appendix~\ref{app:per-type-analysis}.

\paragraph{Impact of the \sys tokenizer.}
\label{sec:exp_tokenizer}

As shown in Table~\ref{tab:main-indomain-results}, replacing the BPE tokenizer with the \sys tokenizer slightly decreases recall in some settings. However, it substantially improves precision, with gains of approximately 18--20 percentage points in W. Strict Typed precision across model sizes. The gain is slightly larger under Strict evaluation than under Partial evaluation, indicating that the tokenizer is especially effective in improving boundary-complete span recovery. This suggests that the \sys tokenizer helps the model de-identify only the actual PII spans, thereby reducing unnecessary masking and minimizing information loss in the original documents.

This precision improvement can be explained by clean PII-span coverage at the tokenizer level. When token boundaries do not align with PII span boundaries, a token may contain both PII and surrounding non-PII text, forcing the model to mask more text than necessary. To examine this effect, we tokenize the judgment documents and measure both clean PII-span coverage and token count. The results are summarized in Table~\ref{tab:tokenizer-boundary-analysis}. The BPE tokenizer cleanly covers only 70.8\% of gold PII spans without extra non-space characters, meaning that many PII spans cannot be precisely isolated at the token level. 

Thunder-DeID addresses this issue by using the Mecab-ko morphological analyzer before BPE, improving clean coverage to 92.4\%, but still leaving 7.6\% of gold PII spans with extra non-space characters. In contrast, the \sys tokenizer achieves 99.7\% clean PII-span coverage, indicating that it can almost always produce token boundaries that match PII spans. This improved clean coverage directly supports the precision gains observed above. 

Clean PII-span coverage can be improved by finer-grained tokenization, such as splitting text at the character level, since finer boundaries make it easier to isolate exact PII spans. However, this substantially increases the number of tokens, reducing the maximum document length that can be processed at once and slowing inference. 
% The \sys tokenizer achieves this balance: while \hl{Unicode character-level tokenization achieves 100.0\% clean PII-span coverage but requires 1,000 tokens per 1,000 input characters---2.08 times as many as \sys \ ---and} Mecab-ko+BPE increases the token count by 17\%, the \sys tokenizer increases it by only 5\%, while still achieving near-perfect clean PII-span coverage.
The \sys tokenizer achieves this balance by nearly matching character-level tokenization in clean coverage (99.7\% vs. 100.0\%) while using fewer than half as many tokens (480.2 vs. 1,000 per 1,000 characters). Its token count remains close to that of standard BPE (459.5) and lower than those of both Mecab-ko+BPE variants (509.0--539.1).

\begin{table}[t]
\centering
\small
\setlength{\tabcolsep}{4pt}
\renewcommand{\arraystretch}{1.05}
\resizebox{\columnwidth}{!}{%
\begin{tabular}{llrrrrr}
\toprule
\textbf{Model} & \textbf{Tokenizer} & \textbf{0} & \textbf{50} & \textbf{100} & \textbf{200} & \textbf{1{,}000} \\
\midrule
150M & BPE        & 0.407 & 0.647 & 0.689 & 0.761 & 0.876 \\
150M & \sys tok.  & 0.453 & 0.679 & 0.727 & 0.790 & 0.894 \\
395M & BPE        & 0.448 & 0.736 & 0.809 & 0.859 & 0.888 \\
395M & \sys tok.  & 0.491 & 0.742 & 0.818 & 0.879 & 0.907 \\
750M & BPE        & 0.451 & 0.787 & 0.869 & 0.888 & 0.907 \\
750M & \sys tok.  & 0.485 & \textbf{0.814} & 0.889 & 0.908 & 0.930 \\
1B   & BPE        & 0.462 & 0.803 & 0.877 & 0.892 & 0.911 \\
1B   & \sys tok.  & \textbf{0.494} & \textbf{0.814} & \textbf{0.891} & \textbf{0.909} & \textbf{0.933} \\
\bottomrule
\end{tabular}%
}
\caption{Data-efficient adaptation from court judgments to general sensitive documents. Values are W. Strict Typed F1-score at representative training sizes; column 0 is zero-shot. Best values in each column are shown in bold. Full F1-score/P/R results over all subset sizes are reported in Appendix~\ref{app:full_results}.}
\label{tab:gendoc-typed-compact}
\vspace{-8pt}
\end{table}

\paragraph{Broader sensitive documents.}

We next examine whether judgment-trained \sys checkpoints can transfer to broader sensitive documents. To this end, we evaluate them on the general-document benchmark, a non-judgment Korean document benchmark described in Appendix~\ref{app:general_docs}. For evaluation, models are initialized from checkpoints trained on the judgment corpus and are first evaluated on the general-document benchmark before any fine-tuning on general-document examples. They are then fine-tuned on progressively larger subsets of the general-document training set, ranging from 25 to 1,000 examples, while the validation and test sets are kept fixed.

The general-document benchmark is annotated with the same PII taxonomy used for judgments, but it contains only 18 PII types. Therefore, evaluation is restricted to these labels, and predictions for all other labels are treated as non-PII. This setup allows us to measure how effectively judgment-trained \sys adapts to broader sensitive documents with only a small number of general-document training examples.

The results are summarized in Table~\ref{tab:gendoc-typed-compact}. Zero-shot domain transfer is challenging, with W. Strict Typed F1-scores ranging from 0.41 to 0.49. However, fine-tuning on a small number of general-document examples substantially improves performance. With only 50 general-document examples, \sys reaches 0.68--0.81 W. Strict Typed F1-score, and with 1,000 examples, it achieves 0.89--0.93. The \sys tokenizer consistently outperforms the BPE tokenizer across all model sizes and training subset sizes, with the best performance achieved by the 1B model, reaching 0.933 W. Strict Typed F1-score. These results show that judgment-trained \sys can be adapted to broader sensitive-document settings using a relatively small number of annotated general-document examples. 
% These results show that judgment-trained \sys can be effectively transferred to broader sensitive-document settings with relatively small amounts of additional annotation.

We additionally compare \sys with three general-purpose PII systems on the same 1,000-document general-document test set: OpenAI Privacy Filter~\cite{debourcy2026modelcardopenaiprivacy}, Korean-PII-Masking-BERT~\cite{lee2025privacy}, and Microsoft Presidio~\cite{MsPresidio}. Because their native label inventories differ from ours, we collapse all predicted PII categories into binary PII/non-PII predictions and evaluate them using W. Strict Binary precision, recall, and F1. The results are shown in Table~\ref{tab:general-pii-baselines}.

% \begin{table}[t]
% \centering
% \footnotesize
% \renewcommand{\arraystretch}{1.05}
% \begin{tabular*}{\columnwidth}{@{\extracolsep{\fill}}lrrr@{}}
% \toprule
% \multirow{2}{*}{\textbf{Model}}
% & \multicolumn{3}{c}{\textbf{W. Strict Binary}} \\
% \cmidrule(lr){2-4}
% & \textbf{F1} & \textbf{P} & \textbf{R} \\
% \midrule
% OpenAI Privacy Filter
% & 0.508 & 0.463 & 0.563 \\
% Korean-PII-Masking-BERT
% & 0.427 & \textbf{0.793} & 0.292 \\
% Microsoft Presidio
% & 0.171 & 0.115 & 0.339 \\
% \midrule
% Thunder-DeID+(150M)
% & \textbf{0.602} & 0.544 & \textbf{0.674} \\
% \bottomrule
% \end{tabular*}
% \caption{Performance of general-purpose PII systems and the zero-shot 150M \sys model on the 1,000-document general-document test set under W. Strict Binary evaluation. The \sys model uses the proposed tokenizer and is evaluated without fine-tuning on general-document data.}
% \label{tab:general-pii-baselines}
% \vspace{-6pt}
% \end{table}

\begin{table}[t]
\centering
\footnotesize
\renewcommand{\arraystretch}{1.05}
\begin{tabular*}{\columnwidth}{@{\extracolsep{\fill}}p{0.46\columnwidth}rrr@{}}
\toprule
\multirow{2}{*}{\textbf{Model}}
& \multicolumn{3}{c}{\textbf{W. Strict Binary}} \\
\cmidrule(lr){2-4}
& \textbf{F1} & \textbf{P} & \textbf{R} \\
\midrule
OpenAI Privacy Filter
& 0.508 & 0.463 & 0.563 \\
Korean-PII-Masking-BERT
& 0.427 & \textbf{0.793} & 0.292 \\
Microsoft Presidio
& 0.171 & 0.115 & 0.339 \\
\midrule
Thunder-DeID+(150M)
& \textbf{0.602} & 0.544 & \textbf{0.674} \\
\bottomrule
\end{tabular*}
\caption{Performance of general-purpose PII systems and the zero-shot 150M \sys model on the 1,000-document general-document test set under W. Strict Binary evaluation. The \sys model uses the proposed tokenizer and is evaluated without fine-tuning on general-document data.}
\label{tab:general-pii-baselines}
\vspace{-6pt}
\end{table}

OpenAI Privacy Filter performs best among the three external baselines, achieving 0.508 W. Strict Binary F1. In comparison, even the smallest 150M \sys model achieves 0.602 without fine-tuning on general-document data. This suggests that PII detection learned from court judgments generalizes beyond the judgment domain even before adaptation to general documents. Full model-size results under W. Strict Binary evaluation are reported in Appendix~\ref{app:full_results}.

\paragraph{Impact of the refined PII taxonomy.}
To examine whether the refined PII taxonomy is associated with improved de-identification performance, we compare fine-tuning under the Thunder-DeID data and annotation scheme with fine-tuning under the \sys data and annotation scheme. For each model size, both experiments start from the Thunder-DeID pretrained weights and use the Mecab-ko+BPE tokenizer with the same optimization settings. Thunder-DeID does not include intellectual-property judgments, so the 100 such documents in the \sys corpus are excluded from this comparison. We then construct the Thunder-DeID comparison set to closely match the remaining \sys corpus in size and category composition while retaining the existing train, validation, and test partitions of both corpora. This yields training sets of 4,620 documents (2,367 civil, 2,093 criminal, and 160 administrative) for Thunder-DeID and 4,619 documents (2,361 civil, 2,091 criminal, and 167 administrative) for \sys.

Models trained under the two data and annotation schemes are evaluated on a 574-document judgment test set (283 civil, 277 criminal, and 14 administrative). For each document, the underlying PII mentions designated for de-identification are held fixed, while two gold annotation versions are prepared separately under the Thunder-DeID and \sys taxonomies and annotation rules. Each model is evaluated against the annotations corresponding to its data and annotation scheme. We report W. Strict Binary F1 for type-agnostic PII detection and W. Strict Typed F1 as a taxonomy-specific diagnostic. The results are summarized in Table~\ref{tab:taxonomy_matched}.

% \begin{table}[t]
% \centering
% \footnotesize
% \setlength{\tabcolsep}{2.5pt}
% \renewcommand{\arraystretch}{1.08}
% \begin{tabular*}{\columnwidth}{@{\extracolsep{\fill}}lccc@{}}
% \toprule
% & & \multicolumn{2}{c}{\textbf{Data / annotation}} \\[-3pt]
% \cmidrule(lr){3-4}
% \textbf{Metric}
% & \shortstack{\textbf{Backbone}\\\textbf{(Thunder-DeID)}}
% & \shortstack{\textbf{Thunder-}\\\textbf{DeID}}
% & \shortstack{\textbf{Thunder-}\\\textbf{DeID+}} \\
% \midrule
% \multirow{3}{*}{\shortstack[l]{W. Strict\\Binary F1}}
% & 370M & 0.612 & 0.755 \\
% & 800M & 0.608 & 0.764 \\
% & 1.5B & 0.617 & 0.736 \\
% \midrule
% \multirow{3}{*}{\shortstack[l]{W. Strict\\Typed F1}}
% & 370M & 0.476 & 0.722 \\
% & 800M & 0.452 & 0.724 \\
% & 1.5B & 0.474 & 0.691 \\
% \bottomrule
% \end{tabular*}
% \caption{Comparison of fine-tuning with the Thunder-DeID and \sys data and annotation schemes using three Thunder-DeID pretrained backbones. W. Strict Binary F1 provides a type-agnostic comparison across the two schemes. W. Strict Typed F1 is computed within each taxonomy's label space and is reported as a taxonomy-specific diagnostic. All results are evaluated on the 574-document judgment test set.}
% \label{tab:taxonomy_matched}
% \vspace{-8pt}
% \end{table}

\begin{table}[t]
\centering
\fontsize{7.5}{8.5}\selectfont
\setlength{\tabcolsep}{2.2pt}
\renewcommand{\arraystretch}{1.04}
\begin{tabular*}{\columnwidth}{@{\extracolsep{\fill}}lccc@{}}
\toprule
& & \multicolumn{2}{c}{\textbf{Data / annotation}} \\[-3pt]
\cmidrule(lr){3-4}
\textbf{Metric}
& \shortstack{\textbf{Backbone}\\\textbf{(Thunder-DeID)}}
& \shortstack{\textbf{Thunder-}\\\textbf{DeID}}
& \shortstack{\textbf{Thunder-}\\\textbf{DeID+}} \\
\midrule
\multirow{3}{*}{\shortstack[l]{W. Strict\\Binary F1}}
& 370M & 0.612 & 0.755 \\
& 800M & 0.608 & 0.764 \\
& 1.5B & 0.617 & 0.736 \\
\midrule
\multirow{3}{*}{\shortstack[l]{W. Strict\\Typed F1}}
& 370M & 0.476 & 0.722 \\
& 800M & 0.452 & 0.724 \\
& 1.5B & 0.474 & 0.691 \\
\bottomrule
\end{tabular*}
\caption{Comparison of fine-tuning with the Thunder-DeID and \sys data and annotation schemes using three Thunder-DeID pretrained backbones. W. Strict Binary F1 provides a type-agnostic comparison across the two schemes. W. Strict Typed F1 is computed within each taxonomy's label space and is reported as a taxonomy-specific diagnostic. All results are evaluated on the 574-document judgment test set.}
\label{tab:taxonomy_matched}
\vspace{-8pt}
\end{table}

Fine-tuning with the \sys data and annotation scheme yields W. Strict Binary F1 gains of 0.120--0.156 over the Thunder-DeID scheme across all three pretrained backbones. By holding the pretrained weights, tokenizer, optimization settings, and evaluation documents fixed, we reduce several major confounding factors in the system-level comparison. Because W. Strict Binary F1 ignores PII type, these results show that the performance difference persists even when type distinctions are removed from the evaluation, indicating an advantage in detecting complete PII word-segments.

W. Strict Typed F1 is computed under different label spaces, so the absolute scores are not directly comparable across the two taxonomies. Nevertheless, the results reflect the motivation for refining the taxonomy. In Thunder-DeID, the same entity may fall under multiple labels; for example, a company can be described as both "중공업회사" (heavy industry company) and "주식회사" (corporation). This overlap makes the intended prediction target ambiguous. The refined taxonomy largely eliminates such overlap and organizes labels at more consistent levels of granularity, providing clearer prediction targets. The training corpora and annotation rules are not identical, so the gains cannot be attributed to the taxonomy alone. Taken together, however, the binary and typed results suggest that the refined taxonomy and annotation design help explain the performance gains.

    \section{Conclusion}
\label{sec:conclusion}

\sys{} is a de-identification framework for Korean court judgments that integrates tokenization, label design, and evaluation. The framework refines the personally identifiable information (PII) taxonomy into an extensible three-tier scheme that addresses the granularity and redundancy issues present in previous Korean legal PII labels. It introduces a boundary-aware tokenizer that augments pretrained BPE with N-gram and boundary-scoring components, thereby improving alignment between token boundaries and PII span boundaries, even in the presence of Korean particles and endings. Additionally, the framework adopts a tokenizer-independent word-segment-level evaluation protocol using Strict/Partial and Typed/Binary F1-scores to enable fair model comparison. Experimental results demonstrate that the framework establishes a robust foundation for Korean de-identification, achieving effective performance on both court judgments and non-judgment documents. 

    \section*{Limitations}

The main judgment experiments use surrogate-realized documents, not raw judgments. This choice is necessary because most publicly released Korean judgments contain anonymization placeholders, and replacement lets us evaluate realistic PII surface forms while preserving the original legal context. The results should therefore be interpreted within this controlled surrogate setting. Our evaluation measures span-level de-identification performance and does not directly assess document-level re-identification risk.

The general-document benchmark includes both human-curated and synthetic documents. Synthetic data help cover practical document formats and rare PII configurations, but they may not fully capture the diversity, noise, and ambiguity of real documents. The cross-domain results should therefore be read as evidence of adaptation potential, not as exhaustive coverage of all non-judgment settings.

Our method and evaluation are designed for Korean. In particular, the word-segment metric relies on whitespace-delimited units that largely correspond to Korean eojeols. Applying the same evaluation protocol to languages without explicit whitespace would require additional segmentation choices.

    \section*{Ethical Considerations}

% All court judgment  were lawfully collected and processed pursuant to Article 24-2 of the Korean Copyright Act, Free Use of Public Works. 

All court judgments used in this work were collected and processed from publicly available sources pursuant to Article 24-2 of the Korean Copyright Act~\citep{korea_copyright}, which permits free use of public works. Because court judgments may still contain sensitive information, we used the corpus only for research on de-identification and did not attempt to re-identify any individual or party.

For model training and evaluation, annotated PII spans were replaced with type-compatible surrogate strings. Structured personal identifiers, such as names, phone numbers, registration numbers, account numbers, and online identifiers, were generated synthetically and do not correspond to real individuals. Some surrogate values for institutions, businesses, organizations, products, and place names were sampled from public resources; these are used only to construct realistic surface forms for de-identification research.

The general-document benchmark contains both human-curated and synthetic documents. Synthetic documents were generated to cover document formats and PII configurations that are difficult to collect at scale without exposing private data. The benchmark is intended for evaluating de-identification methods, not for profiling, identifying, or making decisions about individuals.

% De-identification is a privacy-sensitive task. Although the proposed model improves detection and boundary precision, automatic predictions may still miss sensitive information or mask text that should remain visible. Any real-world deployment should include human review, institution-specific masking policies, and audit procedures.
    \section*{Acknowledgments}

This work was partially supported by the National Research Foundation of Korea (NRF) under Grant No. RS-2023-00222663 (Center for Optimizing Hyperscale AI Models and Platforms) and under Grant No. A400-20260031, and by the Institute for Information and Communications Technology Promotion (IITP) under Grant No. 2018-0-00581 (CUDA Programming Environment for FPGA Clusters) and No. RS-2025-02304554 (Efficient and Scalable Framework for AI Heterogeneous Cluster Systems), all funded by the Ministry of Science and ICT (MSIT) of Korea. It was also partially supported by the Korea Health Industry Development Institute (KHIDI) under Grant No. RS-2025-25454559 (Frailty Risk Assessment and Intervention Leveraging Multimodal Intelligence for Networked Deployment in Community Care), funded by the Ministry of Health and Welfare (MOHW) of Korea. Additional support was provided by the BK21 Plus Program for Innovative Data Science Talent Education (Department of Data Science, Seoul National University, No. 5199990914569) and the BK21 FOUR Program for Intelligent Computing (Department of Computer Science and Engineering, Seoul National University, No. 4199990214639), both funded by the Ministry of Education (MOE) of Korea. This work was also partially supported by the Advanced GPU Utilization Support Program, funded by the Ministry of Science and ICT (MSIT) of Korea and operated by the National IT Industry Promotion Agency (NIPA), as well as by MSIT through NIPA as part of the Sovereign AI Foundation Model Project. It was also carried out under the project's Data Track, led by MSIT and supported by the National Information Society Agency (NIA). Research facilities were provided by the Institute of Computer Technology (ICT) at Seoul National University.

    % \clearpage
    \bibliography{custom}

    \clearpage
    \appendix
    \section*{\centering\textbf{Appendix}}

\section{Details in Corpus Construction}
\label{app:corpus_construction}

% ===== OCR ====================================

\subsection{OCR}
\label{app:ocr}
\paragraph{OCR procedure.}
For judgments available only as scanned image documents without embedded text, we performed OCR using Gemini 2.5 Flash~\cite{gemini25}. Just under 10\% of the documents required OCR. Table~\ref{tab:ocr-config} summarizes the OCR configuration. 

\begin{table}[htbp]
\centering
\footnotesize
\setlength{\tabcolsep}{4pt}
\renewcommand{\arraystretch}{1}
    \begin{tabular}{@{}p{0.18\columnwidth}|p{0.72\columnwidth}@{}}
    
    \toprule
    Model & \texttt{gemini-2.5-flash} \\ 
    
    \midrule
    % \begin{minipage}[t]{\linewidth}
    %     System\\
    %     prompt
    % \end{minipage}
    % & None \\ 
    
    % \midrule
    \begin{minipage}[t]{\linewidth}
        User\\
        prompt\\
        (original)
    \end{minipage}
    & 이 판결문 PDF의 모든 내용을 텍스트로 변환해줘. 내용은 그대로 유지하고 (한자 및 숫자 포함) 줄바꿈도 살려줘. \\ 

    \midrule
    \begin{minipage}[t]{\linewidth}
        User\\
        prompt\\
        (English\\
        translation)
    \end{minipage}
    & Convert all contents of this judgment PDF into text. Preserve the content as-is, including Hanja characters and numbers, and preserve line breaks. \\

    \bottomrule
    \end{tabular}
\caption{OCR configuration used for scanned PDFs.
\label{tab:ocr-config}}
\end{table}

\newcommand{\annred}[1]{%
  {\color{red}%\bfseries
  <\kern0pt<\kern0pt<#1>\kern0pt>\kern0pt>}%
}

\begin{table*}[!t]
\centering
\setlength{\tabcolsep}{6pt}
\renewcommand{\arraystretch}{1.15}

{\scriptsize
\begin{tabular}{@{}c|>{\raggedright\arraybackslash}p{0.45\textwidth}|>{\raggedright\arraybackslash}p{0.45\textwidth}@{}}
\toprule
\small\textbf{Line} & \small\textbf{Source judgment text} & \small\textbf{Annotated text} \\

\midrule
1 &
【피 고 인】 &
【피 고 인】 \annred{피고인:인명} \\

2 &
【검\hspace*{2em}사】 김유나(기소), 성기범(공판) &
【검\hspace*{2em}사】 김유나(기소), 성기범(공판) \\

3 &
【주\hspace*{2em}문】 &
【주\hspace*{2em}문】 \\

4 &
피고인을 징역 3월에 처한다. &
피고인을 징역 3월에 처한다. \\

5 &
&
\\

6 &
【이\hspace*{2em}유】【범죄사실】\hspace*{1.5em}피고인은, 사실 자신 명의의 재산이 없고 그가 운영하는 `○○○'이라는 상호의 금은방에 시가 합계 약 2억 원 상당의 귀금속을 보유하고 있었으나 새마을금고에 대한 대출금 채무가 10억여 원에 달하고 사채 6,000만 원 상당을 월 10\%의 이자로 빌린 상황이어서 그 이자도 제대로 내기 어려웠으므로 피해자로부터 돈을 빌리더라도 이를 변제할 의사나 능력이 없었음에도, &
【이\hspace*{2em}유】【범죄사실】 피고인은, 사실 자신 명의의 재산이 없고 그가 운영하는 `\annred{○○○:공산품판매업}'이라는 상호의 금은방에 시가 합계 약 2억 원 상당의 귀금속을 보유하고 있었으나 \annred{새마을금고:은행업}에 대한 대출금 채무가 10억여 원에 달하고 사채 6,000만 원 상당을 월 10\%의 이자로 빌린 상황이어서 그 이자도 제대로 내기 어려웠으므로 피해자로부터 돈을 빌리더라도 이를 변제할 의사나 능력이 없었음에도, \\

7 &
\hspace*{1.5em}1. 2009. 12. 14.경 청주시 (주소 1 생략) 소재 위 `○○○' 금은방에서, 피해자 공소외 2에게 ``금을 매입하려고 하는데 자금이 모자라니 이를 빌려주면 2010. 3. 말까지 갚겠다''고 거짓말하여, 이에 속은 피해자로부터 차용금 명목으로 2009. 12. 15. 자신의 제일은행 계좌(계좌번호 생략)로 1,500만 원을 송금 받았고, &
\hspace*{1.5em}1. 2009. 12. 14.경 청주시 \annred{(주소 1 생략):시아래주소} 소재 위 `\annred{○○○:공산품판매업}' 금은방에서, 피해자 공소외 \annred{2:인명}에게 ``금을 매입하려고 하는데 자금이 모자라니 이를 빌려주면 2010. 3. 말까지 갚겠다''고 거짓말하여, 이에 속은 피해자로부터 차용금 명목으로 2009. 12. 15. 자신의 \annred{제일:은행업}은행 계좌\annred{(계좌번호 생략):계좌번호}로 1,500만 원을 송금 받았고, \\

8 &
\hspace*{1.5em}2. 2009. 12. 23. 전항과 같은 장소에서 피해자에게 ``금값이 올랐으니 금을 사두면 돈이 된다, 금 구입비용을 빌려주면 이자와 함께 3개월 후에 변제하겠다''고 거짓말하여, 이에 속은 피해자로부터 차용금 명목으로 전항 기재와 같은 계좌로 1,500만 원을 송금 받았고, &
\hspace*{1.5em}2. 2009. 12. 23. 전항과 같은 장소에서 피해자에게 ``금값이 올랐으니 금을 사두면 돈이 된다, 금 구입비용을 빌려주면 이자와 함께 3개월 후에 변제하겠다''고 거짓말하여, 이에 속은 피해자로부터 차용금 명목으로 전항 기재와 같은 계좌로 1,500만 원을 송금 받았고, \\

9 &
\hspace*{1.5em}3. 2010. 2. 20. 위 1.항과 같은 장소에서 피해자에게 ``금값이 급등하여 금을 사 두어야 하는데 금 구입자금이 필요하다, 금 구입비용을 빌려주면 이자와 함께 3개월 후에 변제하겠다''고 거짓말하여, 이에 속은 피해자로부터 차용금 명목으로 위 1.항 기재와 같은 계좌로 1,000만 원을 송금 받는 등 &
\hspace*{1.5em}3. 2010. 2. 20. 위 1.항과 같은 장소에서 피해자에게 ``금값이 급등하여 금을 사 두어야 하는데 금 구입자금이 필요하다, 금 구입비용을 빌려주면 이자와 함께 3개월 후에 변제하겠다''고 거짓말하여, 이에 속은 피해자로부터 차용금 명목으로 위 1.항 기재와 같은 계좌로 1,000만 원을 송금 받는 등 \\

10 &
3회에 걸쳐 피해자를 기망하여 피해자로부터 합계 4,000만 원을 교부받아 이를 편취하였다. &
3회에 걸쳐 피해자를 기망하여 피해자로부터 합계 4,000만 원을 교부받아 이를 편취하였다. \\

11 &
【증거의 요지】\hspace*{1.5em}1. 피고인의 법정진술 &
【증거의 요지】\hspace*{1.5em}1. 피고인의 법정진술 \\

12 &
\hspace*{1.5em}1. 공소외 2에 대한 경찰진술조서 &
\hspace*{1.5em}1. 공소외 \annred{2:인명}에 대한 경찰진술조서 \\

13 &
\hspace*{1.5em}1. 이체확인증사본, 재산조회회보서 &
\hspace*{1.5em}1. 이체확인증사본, 재산조회회보서 \\

14 &
【법령의 적용】\hspace*{1.5em}1. 범죄사실에 대한 해당법조 및 형의 선택 &
【법령의 적용】\hspace*{1.5em}1. 범죄사실에 대한 해당법조 및 형의 선택 \\

15 &
형법 제347조 제1항(판시 각 사기의 점을 포괄하여), 징역형 선택 &
형법 제347조 제1항(판시 각 사기의 점을 포괄하여), 징역형 선택 \\

16 &
【양형의 이유】\hspace*{1em}피고인이 이 사건 범행을 자백하고 있고 범죄 전력이 전혀 없는 초범임을 감안하더라도, 편취 금액이 4,000만 원 상당에 이르러 결코 적지 않은 금액임에도 피해 변제를 위한 아무런 노력도 기울이지 않고 있는 점에서 피해자가 느꼈을 정신적·경제적 고통 등을 고려하면 피고인에 대하여는 실형의 선고가 불가피하다. 다만 기록에 나타난 형법 제51조 소정의 제반 양형의 요소를 두루 참작하여, 주문과 같은 형을 정한다. &
【양형의 이유】\hspace*{1em}피고인이 이 사건 범행을 자백하고 있고 범죄 전력이 전혀 없는 초범임을 감안하더라도, 편취 금액이 4,000만 원 상당에 이르러 결코 적지 않은 금액임에도 피해 변제를 위한 아무런 노력도 기울이지 않고 있는 점에서 피해자가 느꼈을 정신적·경제적 고통 등을 고려하면 피고인에 대하여는 실형의 선고가 불가피하다. 다만 기록에 나타난 형법 제51조 소정의 제반 양형의 요소를 두루 참작하여, 주문과 같은 형을 정한다. \\

17 &
&
\\

18 &
판사 윤성묵 &
판사 윤성묵\\

\bottomrule
\end{tabular}
}

\caption{Example of full-context annotation. The source judgment text is preserved, while inline markers are inserted around spans to be de-identified. Line numbers are added for side-by-side comparison, and line breaks may be adjusted for table layout.}
\label{tab:annotation-examples}
\end{table*}

\begin{table*}[!t]
\centering
\footnotesize
\setlength{\tabcolsep}{4pt}
\renewcommand{\arraystretch}{1}
    \begin{tabular}{@{}p{0.18\textwidth}|p{0.72\textwidth}@{}}
    \toprule
    Model & Qwen3.6-35B-A3B-UD-Q4\_K\_M.gguf \\ 
        
    \midrule
        \begin{minipage}[t]{\linewidth}
            System\\
            prompt\\
            (original)
        \end{minipage}
    & 
        \begin{minipage}[t]{\linewidth}
            당신은 한국어 판결서 OCR 후처리기입니다.\\
            입력 텍스트의 띄어쓰기와 줄바꿈만 교정하세요.\\
            반드시 지킬 규칙:\\
            - 글자, 숫자, 기호, 한자, 영문, 마크다운 기호를 절대 바꾸지 마세요.\\
            - 익명화 태그 $\lll$...$\ggg$ 안의 내용과 태그 기호를 절대 바꾸지 마세요.\\
            - 허용되는 변경은 스페이스, 탭, 줄바꿈의 추가/삭제/변경뿐입니다.\\
            - 문장을 요약하거나 표현을 바꾸거나 맞춤법을 고치지 마세요.\\
            - 사고 과정, 설명, 분석, 주석을 출력하지 마세요.\\
            - 출력에는 교정된 텍스트만 포함하세요.
        \end{minipage} \\ 
        
    \midrule
        \begin{minipage}[t]{\linewidth}
            System\\
            prompt\\
            (English\\
            translation)
        \end{minipage}
    & 
        \begin{minipage}[t]{\linewidth}
            You are a Korean court judgment OCR post-processor.\\
            Correct only the spacing and line breaks in the input text.\\
            Rules that must be followed:\\
            - Never change letters, numbers, symbols, Chinese characters, English text, or markdown symbols.\\
            - Never change the content inside anonymization tags $\lll$...$\ggg$ or the tag symbols themselves.\\
            - The only permitted changes are adding, deleting, or modifying spaces, tabs, and line breaks.\\
            - Do not summarize sentences, rephrase them, or correct spelling.\\
            - Do not output reasoning, explanations, analysis, or comments.\\
            - Include only the corrected text in the output.
        \end{minipage} \\ 
        
    \midrule
        \begin{minipage}[t]{\linewidth}
            User\\
            prompt\\
            (original)
        \end{minipage}
    & 
        \begin{minipage}[t]{\linewidth}
            아래 텍스트에서 띄어쓰기와 줄바꿈만 정상화하세요.\\
            공백을 제외한 모든 문자는 입력과 정확히 같아야 합니다.\\
            \\
            {[}입력 시작{]}\\
            \texttt{\{annotated text\}}\\
            {[}입력 끝{]}
        \end{minipage} \\ 
        
    \midrule
        \begin{minipage}[t]{\linewidth}
            User\\
            prompt\\
            (English\\
            translation)
        \end{minipage}
    & 
        \begin{minipage}[t]{\linewidth}
            Normalize only the spacing and line breaks in the text below.\\
            All non-whitespace characters must be exactly the same as in the input.\\
            \\
            {[}Input start{]}\\
            \texttt{\{annotated text\}}\\
            {[}Input end{]}
        \end{minipage} \\ 
    
    \bottomrule
    \end{tabular}
\caption{Line-break normalization configuration.
\label{tab:normalization-config}}
\end{table*}

\paragraph{Boilerplate removal.} 
The OCR outputs occasionally included boilerplate text corresponding to document headers, footers, page markers, or other layout artifacts. We removed these artifacts using rule-based regular expressions. The removed boilerplate mainly included court-logo artifacts, such as \texttt{대한민국법원}, and their OCR-corrupted variants; page markers, such as \texttt{-1-}; repeated legal notices, such as \texttt{본 판결문은 판결서 인터넷열람 사이트에서 열람·출력되었습니다.}; and date-stamped metadata, such as \texttt{비실명처리일자: YYYY-MM-DD}. The rules were applied only to predefined boilerplate patterns and were not intended to modify the substantive judgment text.

% ===== Annotation ====================================

\subsection{Annotation}
\label{app:annotation}
\paragraph{Annotation procedure and quality control.}
Annotation was conducted by 9 members of the author team, all of whom are fluent in Korean and familiar with NLP annotation tasks. 

Since the corpus contains both anonymized and partially non-anonymized judgments, the task was not limited to labeling existing anonymization placeholders, i.e. tokens such as ``○○'', ``A'' or ``1'' that already replace de-identified information in court-released judgments. Instead, annotators examined each judgment in full and additionally marked spans that should be de-identified according to the annotation rules\footnote{Korean Supreme Court Regulation No. 2809 and Judicial Rule No. 1778}. Newly identified spans accounted for 15.01\% of the final annotations, showing that full-document review was necessary. 

This setting required explicit guidelines because the same surface expression may or may not require masking depending on its role in the case narrative and its potential to identify the parties involved.
For example, a personal name is masked when it belongs to a party to the case, but the name of an attorney or legal representative acting in their official capacity is left unmasked.

% To ensure consistency and quality, annotators were trained with guidelines covering Korean court anonymization practices, the newly designed PII taxonomy (\ref{sec:Method:03_sub_PIItaxonomy}), and rules for applying the taxonomy to document-level de-identification. The training emphasized contextual de-identification judgments, span boundary judgments, and ambiguous cases where legal relevance or incident association affects whether a mention should be masked. We then followed an iterative annotation process adapted from prior work~\cite{thunder-deid}: initial rule design, document-level annotation, consistency review, rule refinement, and final adjudication.

To ensure consistency and quality, annotators were trained with guidelines covering Korean court anonymization practices, the newly designed PII taxonomy (Section~\ref{sec:Method:03_sub_PIItaxonomy}), and rules for applying the taxonomy to document-level de-identification. The training emphasized contextual de-identification judgments, span boundary judgments, and ambiguous cases where legal relevance or incident association affects whether a mention should be masked. We then followed an iterative annotation process adapted from prior work~\cite{thunder-deid}: initial rule design, document-level annotation, consistency review, rule refinement, and final adjudication. 
The annotation process consisted of iterative review and adjudication without independent double annotation, and pre-adjudication inter-annotator agreement was therefore not computed. During review, 349 ambiguous or inconsistent annotation issues were recorded, discussed, and incorporated into subsequent guideline updates and final annotations.

% We followed an iterative annotation process adapted from prior
% work~\cite{thunder-deid}: initial rule design, document-level
% annotation, consistency review, rule refinement, and final
% adjudication. The workflow did not use independent double annotation,
% so we do not report a pre-adjudication inter-annotator agreement
% score. During review, 349 ambiguous or inconsistent annotation issues
% were logged and resolved through guideline refinement, cross-checking,
% and final adjudication.

\paragraph{Annotation principles.}
In addition to the PII taxonomy itself, we define context-aware annotation rules that allow selective abstraction and specification. By default, annotators label an entity at the predefined leaf-label level. However, when a more specific attribute is important in the global context of the judgment, the annotator may use a more specific label. Conversely, when the entity cannot be reliably specified at the leaf level, or when the additional specificity is not relevant to the document context, the annotator may assign a more abstract parent label. For example, a person name is labeled with nationality information only when the nationality is explicitly stated or contextually important; otherwise, it is labeled simply as a person name. Likewise, a restaurant mention is labeled with a specific cuisine type only when that subtype is relevant to the case context; otherwise, it is labeled at the general restaurant level.

This context-aware design separates the label used for model prediction from the additional specificity used for augmentation. In particular, specific subtypes can be used to sample more natural replacement entities during augmentation, while the model can still be trained and evaluated at a consistent prediction level. This mitigates the problem in which some examples require only a coarse category prediction whereas others require a fine-grained subtype prediction. As a result, the proposed taxonomy provides a more stable target space for model learning, while retaining enough semantic detail to support realistic document restoration and augmentation.

\paragraph{Annotation format.}
By default, each annotation consists of a mention span and its PII label, written inline as $\lll$MENTION:LABEL$\ggg$. Optional extensions such as $\lll$MENTION:LABEL:SUB-LABEL$\ggg$ were used when selective specification was needed under the annotation principles above.

The marker was applied only to the minimal span that uniquely represents the entity, excluding descriptive prefixes/suffixes or common nouns attached to it. For example, source text spans ``A식당'' (restaurant A) and ``농협은행'' (NongHyup Bank) were annotated as ``$\lll$A:식당$\ggg$식당'' and ``$\lll$농협:은행업$\ggg$은행'', respectively, where ``식당'' denotes a restaurant and ``은행업'' denotes a banking business.

\paragraph{Examples.}
Table~\ref{tab:annotation-examples} presents representative examples of the annotation format and span-boundary decisions.
% Representative annotation examples are shown in Table~\ref{tab:annotation-examples}.

% ===== Normalization ====================================

\subsection{Line-break normalization}
\label{app:normalization}

In some judgments, line wrapping from the original document layout was preserved as newline characters, producing unnatural breaks within sentences or paragraphs. Thus we applied line-break normalization to all annotated judgments, regardless of their source or whether OCR had been performed. 

This step was limited to removing layout-induced newline artifacts, including adjacent spacing artifacts caused by indentation or line wrapping, and was not intended to revise the document text. We corrected these artifacts using a 4-bit quantized version of Qwen3.6-35B-A3B~\cite{qwen36_35b_a3b}. Table~\ref{tab:normalization-config} summarizes the normalization configuration.

Then we verified that the normalization did not paraphrase, rewrite, or otherwise modify the substantive document text.  For each generated chunk, we compared the input and output after removing all whitespace characters and accepted the output only if the resulting strings matched exactly. This ensured that the line-break normalization changed only whitespace and did not introduce any lexical substitutions, spelling corrections, deletion of characters, or changes to anonymization tags. Outputs that failed validation were retried, and persistent failures were manually inspected and corrected.

% ===== Corpus stats. ====================================

\subsection{Corpus statistics and Case-type distributions}
\label{app:corpus_stats}

This appendix provides additional document- and span-level statistics and
case-type distribution details for the annotated judgment corpus. All values are
computed from the raw annotated judgments before replacement.

The 27,992 collected judgments formed the initial document pool, from which approximately 6,000 documents were selected for annotation to cover the major legal areas and 28 case types while avoiding excessive concentration in frequent case types and retaining lower-frequency categories. After annotation and quality filtering, 5,872 documents remained in the final corpus.

\begin{table}[h]
\centering
\small
\setlength{\tabcolsep}{6pt}
\begin{tabular}{lr}
\toprule
Statistic & Value \\
\midrule
Document count & 5,872 \\
Document length, mean & 4,554 chars \\
Document length, median & 3,116 chars \\
Document length, 95th percentile & 12,813 chars \\
\midrule
Annotated spans & 193,004 \\
% Raw inline label strings & 1,231 \\
Spans per document, mean & 32.9 \\
Spans per document, median & 13 \\
\bottomrule
\end{tabular}
\caption{Summary statistics of the final annotated judgment corpus. Document lengths are measured in characters after removing inline annotation markup.}
\label{tab:corpus-summary}
\end{table}

Table~\ref{tab:corpus-summary} summarizes the document- and span-level statistics of the final corpus, including document length and the number of annotated spans per document. The median document contains 13 annotated spans.

\begin{table}[h]
\centering
\small
\setlength{\tabcolsep}{6pt}
\begin{tabular}{lrr}
\toprule
Legal area & Documents & \% \\
\midrule
Civil & 2,958 & 50.4 \\
Criminal & 2,615 & 44.5 \\
Administrative & 199 & 3.4 \\
Intellectual property & 100 & 1.7 \\
\midrule
Total & 5,872 & 100.0 \\
\bottomrule
\end{tabular}
\caption{Legal area distribution of the annotated judgment corpus. Percentages are computed over the 5,872 documents in the final corpus.}
\label{tab:corpus-case-type}
\end{table}

\begin{table*}[t]
\centering
% \footnotesize
\scriptsize
\setlength{\tabcolsep}{8pt}
\renewcommand{\arraystretch}{1.15}
\begin{tabular}{@{}lllrr@{}}
    \toprule
    Legal Area & Case Category & Fine-grained Case Type & Count & Share \\
    
    \midrule
    \multirow{12}{*}{Civil}
        & Compensation for Damages 
            & Compensation for Damages & 750 & 25.4\% \\
        \cmidrule(l){2-5}
        & \multirow{3}{*}{Credit and Debt}
            & Lease & 297 & 10.0\% \\
            & & Payment & 394 & 13.3\% \\
            & & Loan & 370 & 12.5\% \\
        \cmidrule(l){2-5}
        & \multirow{5}{*}{Real Estate}
            & Transfer of Ownership Registration & 210 & 7.1\% \\
            & & Cancellation of Registration & 208 & 7.0\% \\
            & & Delivery of Building & 209 & 7.1\% \\
            & & Delivery of Land & 110 & 3.7\% \\
            & & Partition of Co-owned Property & 110 & 3.7\% \\
        \cmidrule(l){2-5}
        & \multirow{3}{*}{Labor}
            & Wages & 100 & 3.4\% \\
            & & Invalid Dismissal & 100 & 3.4\% \\
            & & Severance Pay & 100 & 3.4\% \\
        \midrule
    \multirow{14}{*}{Criminal}
        & \multirow{2}{*}{Crimes Against the Person}
            & Assault & 200 & 7.6\% \\
            & & Bodily Injury & 198 & 7.6\% \\
        \cmidrule(l){2-5}
            & \multirow{3}{*}{Sexual Offenses}
            & Indecent Act by Compulsion & 193 & 7.4\% \\
            & & Sexual Offenses Against Children and Adolescents & 188 & 7.2\% \\
            & & Sexual Offenses under the Special Act on Sexual Violence & 183 & 7.0\% \\
        \cmidrule(l){2-5}
            & \multirow{4}{*}{Property Crimes}
            & Theft & 200 & 7.6\% \\
            & & Fraud & 249 & 9.5\% \\
            & & Embezzlement & 200 & 7.6\% \\
            & & Breach of Trust & 200 & 7.6\% \\
        \cmidrule(l){2-5}
        & \multirow{5}{*}{Crimes Against Public Safety and Order}
            & Arson & 100 & 3.8\% \\
            & & Accidental Fire & 131 & 5.0\% \\
            & & Gambling & 196 & 7.5\% \\
            & & Narcotics Offenses & 177 & 6.8\% \\
            & & Road Traffic Act Violations & 200 & 7.6\% \\
    \bottomrule
\end{tabular}
\caption{Distribution of the annotated judgment datasets by legal area, case category, and fine-grained case type.}
\label{tab:detailed-case-type-distribution}
\end{table*}

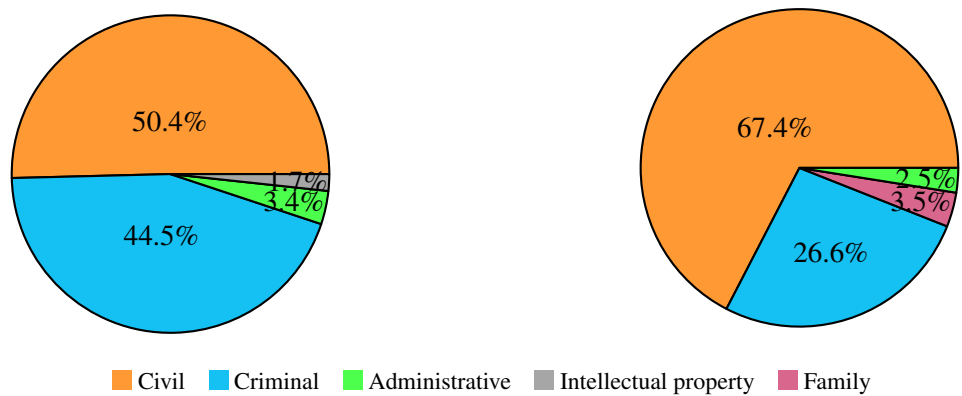
\begin{figure*}[t]
\centering

\begin{minipage}{0.48\textwidth}
\centering
\begin{tikzpicture}
\pie[
    radius=2.1,
    rotate=0,
    color={orange!80, cyan!70, green!70, gray!70}
]{
    50.4/,
    44.5/,
    3.4/,
    1.7/
}

% \node[anchor=east] at (-2.5,0) {
% \footnotesize
% \begin{tabular}{@{}ll@{}}
% \textcolor{orange!80}{\rule{0.8em}{0.8em}} & Civil: 2{,}958 \\
% \textcolor{cyan!70}{\rule{0.8em}{0.8em}} & Criminal: 2{,}615 \\
% \textcolor{green!70}{\rule{0.8em}{0.8em}} & Administrative: 199 \\
% \textcolor{gray!70}{\rule{0.8em}{0.8em}} & Intellectual property: 100
% \end{tabular}
% };
\end{tikzpicture}
\end{minipage}
\hfill
\begin{minipage}{0.48\textwidth}
\centering
\begin{tikzpicture}
\pie[
    radius=2.1,
    rotate=0,
    color={orange!80, cyan!70, purple!60, green!70}
]{
    67.4/,
    26.6/,
    3.5/,
    2.5/
    % 2.5/,
    % 0.1/,
    % 0.0/
}
\end{tikzpicture}
\end{minipage}

\vspace{0.5em}

{\footnotesize
\textcolor{orange!80}{\rule{0.8em}{0.8em}} Civil\quad
\textcolor{cyan!70}{\rule{0.8em}{0.8em}} Criminal\quad
\textcolor{green!70}{\rule{0.8em}{0.8em}} Administrative\quad
\textcolor{gray!70}{\rule{0.8em}{0.8em}} Intellectual property\quad
\textcolor{purple!60}{\rule{0.8em}{0.8em}} Family
}

\caption{Broad-domain distributions of the annotated judgment corpus (left) and court judgments in South Korea (right). Election cases are omitted from the right pie chart because their share is below 0.1\%.}
\label{fig:domain-distribution-comparison}
\end{figure*}

Table~\ref{tab:corpus-case-type} and Table~\ref{tab:detailed-case-type-distribution} report the distribution of the annotated judgment corpus at the legal-area and fine-grained case-type levels. 
The selected annotation subset was constructed to reflect the major composition of Korean court judgments while avoiding excessive concentration in a small number of frequent case types and ensuring coverage of lower-frequency case types. 

Figure~\ref{fig:domain-distribution-comparison}  compares the legal area composition of our annotated corpus with the overall distribution of Korean court judgments. 
This comparison is intended not to reproduce the national distribution exactly, but to show that the corpus preserves the dominant civil and criminal categories while intentionally allocating annotation capacity to less frequent legal areas such as administrative and intellectual property cases. 
Family cases were not included in the collected corpus, as judgments in family litigation are often subject to non-disclosure under Korean law, including Article 10 of the Family Litigation Act.
\FloatBarrier

\section{The Complete PII Taxonomy}
\label{app:PIItaxonomy}

\begin{figure*}[!ht]
    \centering
    \includegraphics[width=1.0\linewidth]{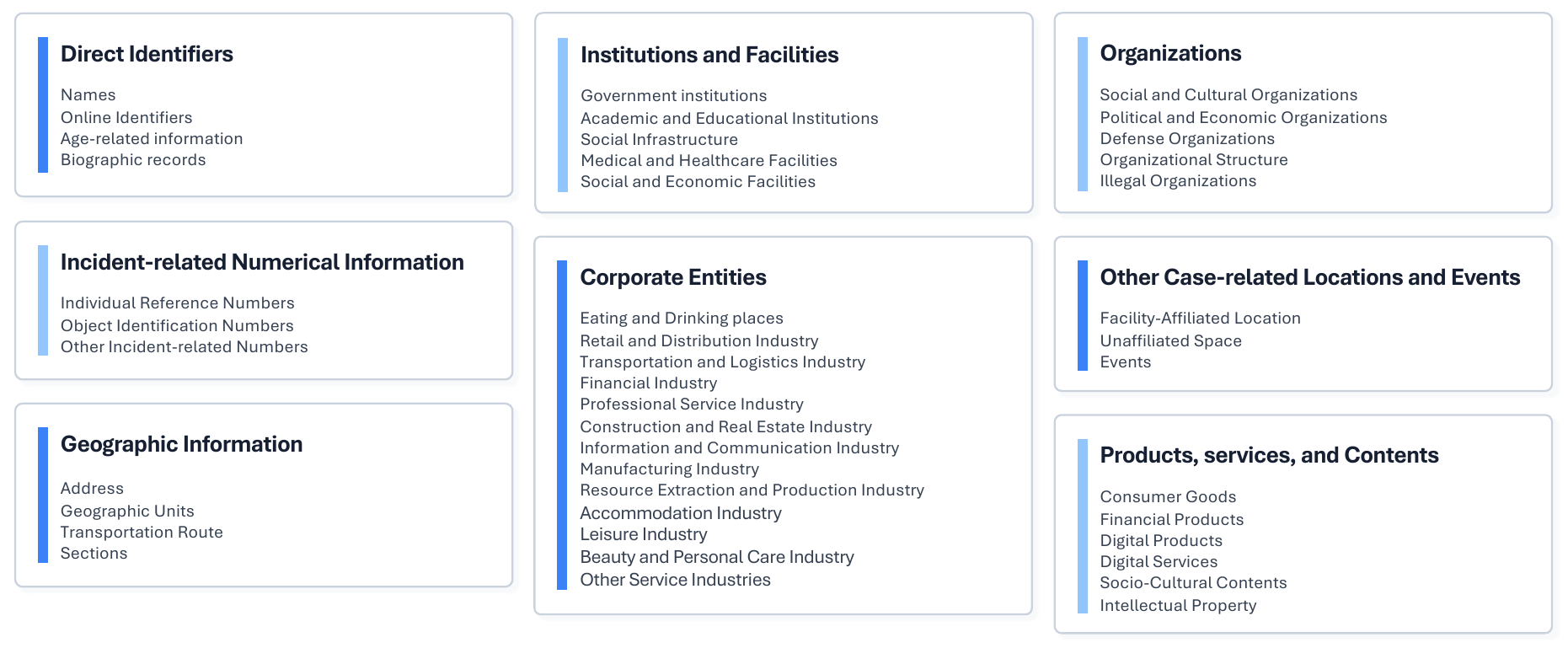}
    \caption{Overview of the refined PII taxonomy. The figure shows the upper two levels of the three-tier hierarchy. The complete label hierarchy is provided in Appendix~\ref{app:PIItaxonomy}.}
    \label{fig:pii-taxonomy-overview}
\end{figure*}

This appendix provides the complete PII taxonomy used in this work. The taxonomy consists of 8 top-level categories, 43 mid-level categories, and 195 fine-grained categories, designed to capture both conventional personal identifiers and context-dependent expressions that may reveal the identities of parties or case-related entities in Korean documents.

\subsection{사건관계인특정정보 (Direct Identifiers)}

\subsubsection{이름 (Names)}
\begin{itemize}
\setlength{\itemsep}{0pt}
\setlength{\parskip}{0pt}
\setlength{\parsep}{0pt}
\setlength{\topsep}{2pt}
\setlength{\partopsep}{0pt}
  \item 인명 (personal names)
  \item 세례명 (baptismal names)
  \item 법명 (Dharma names)
  \item 예명 (stage or pen name)
\end{itemize}

\subsubsection{온라인식별자 (Online Identifiers)}
\begin{itemize}
\setlength{\itemsep}{0pt}
\setlength{\parskip}{0pt}
\setlength{\parsep}{0pt}
\setlength{\topsep}{2pt}
\setlength{\partopsep}{0pt}
  \item 아이디 (ID)
  \item 닉네임 (Nickname)
  \item 이메일주소 (email address)
  \item IP주소 (IP address)
\end{itemize}

\subsubsection{연령정보 (Age-related Information)}
\begin{itemize}
\setlength{\itemsep}{0pt}
\setlength{\parskip}{0pt}
\setlength{\parsep}{0pt}
\setlength{\topsep}{2pt}
\setlength{\partopsep}{0pt}
  \item 출생일자 (date of birth)
  \item 나이 (age)
  \item 여명 (remaining lifespan)
\end{itemize}

\subsubsection{이력사항 (Biographic Records)}
\begin{itemize}
\setlength{\itemsep}{0pt}
\setlength{\parskip}{0pt}
\setlength{\parsep}{0pt}
\setlength{\topsep}{2pt}
\setlength{\partopsep}{0pt}
  \item 학년 (grade level)
  \item 경력 (work experience)
  \item 기수 (cohort)
  \item 수상경력 (awards and honors)
\end{itemize}

\subsection{사건관련숫자정보 (Incident-related
Numerical Information)}

\subsubsection{인적번호 (Individual Reference Numbers)}
\begin{itemize}
\setlength{\itemsep}{0pt}
\setlength{\parskip}{0pt}
\setlength{\parsep}{0pt}
\setlength{\topsep}{2pt}
\setlength{\partopsep}{0pt}
  \item 주민등록번호 (resident registration number)
  \item 계좌번호 (bank account number)
  \item 가상화폐개인지갑주소 (cryptocurrency personal wallet address)
  \item 전화번호 (Phone number)
  \item 군번 (military service number)
  \item 면허번호 (license number)
  \item 사업자등록번호 (business registration number)
  \item 법인등록번호 (corporate registration number)
  \item 여권번호 (passport number)
  \item 훈장번호 (decoration/Medal number)
  \item 수험번호 (examination Registration number)
  \item 보훈번호 (veterans benefits number)
\end{itemize}

\subsubsection{사물·관리 번호 (Object Identification Numbers)}
\begin{itemize}
\setlength{\itemsep}{0pt}
\setlength{\parskip}{0pt}
\setlength{\parsep}{0pt}
\setlength{\topsep}{2pt}
\setlength{\partopsep}{0pt}
  \item 차량번호 (vehicle registration number)
  \item 차대번호 (vehicle identification number, VIN)
  \item 선박번호 (vessel registration number)
  \item 업무관리번호 (task management number)
  \item 문서관리번호 (document management number)
  \item 물류관리번호 (logistics tracking number)
  \item 결제수단관련번호 (payment method number)
  \item 금융증표관련번호 (financial instrument number)
  \item 교통노선관련번호 (transit route number)
  \item 재판사건번호 (court case number)
  \item 경찰사건번호 (police case number)
  \item 비밀번호 (password)
  \item 우편번호 (postal code)
\end{itemize}

\subsubsection{기타사건관련숫자 (Other Incident-related Numbers)}
\begin{itemize}
\setlength{\itemsep}{0pt}
\setlength{\parskip}{0pt}
\setlength{\parsep}{0pt}
\setlength{\topsep}{2pt}
\setlength{\partopsep}{0pt}
  \item 시간정보 (time information)
  \item 운행편번호 (transit service number)
  \item 수치 (numeric figure)
  \item 방번호 (room number)
  \item 출구번호 (exit number)
\end{itemize}

\subsection{지리정보 (Geographic Information)}

\subsubsection{주소 (Address)}
\begin{itemize}
\setlength{\itemsep}{0pt}
\setlength{\parskip}{0pt}
\setlength{\parsep}{0pt}
\setlength{\topsep}{2pt}
\setlength{\partopsep}{0pt}
  \item 행정구역-일반구 (Gu: district-level administrative unit)
  \item 행정구역-동 (Dong: neighborhood-level administrative unit)
  \item 행정구역-읍 (Eup: town-level administrative unit)
  \item 행정구역-면 (Myeon: township-level administrative unit)
  \item 행정구역-리 (Ri: village-level administrative unit)
  \item 국가명 (country name)
  \item 해외지명 (foreign place name)
  \item 해외주소 (foreign address)
  \item 도로명 (road name)
  \item 번지 (lot number)
  \item 건물동호수 (building and unit number)
  \item 건물층 (floor number)
  \item 블록 (block)
  \item 로트 (land lot)
\end{itemize}

\subsubsection{지역명 (Geographic Units)}
\begin{itemize}
\setlength{\itemsep}{0pt}
\setlength{\parskip}{0pt}
\setlength{\parsep}{0pt}
\setlength{\topsep}{2pt}
\setlength{\partopsep}{0pt}
  \item 마을 (town)
  \item 선거구 (voting district)
  \item 지사·지청 (local government office)
  \item 지점 (branch)
  \item 지구 (ward)
  \item 자연지형 (natural landform)
\end{itemize}

\subsubsection{교통로 (Transportation Route)}
\begin{itemize}
\setlength{\itemsep}{0pt}
\setlength{\parskip}{0pt}
\setlength{\parsep}{0pt}
\setlength{\topsep}{2pt}
\setlength{\partopsep}{0pt}
  \item 길 (path)
  \item 거리 (street)
  \item 골목 (alley)
  \item 교차로 (intersection)
  \item 인터체인지 (interchange)
  \item 로터리 (traffic circle)
  \item 도로 (road)
  \item 고속도로 (highway)
\end{itemize}

\subsubsection{구간 (Sections)}
\begin{itemize}
\setlength{\itemsep}{0pt}
\setlength{\parskip}{0pt}
\setlength{\parsep}{0pt}
\setlength{\topsep}{2pt}
\setlength{\partopsep}{0pt}
  \item 고속도로구간 (highway segment)
  \item 철도구간 (railway segment)
\end{itemize}

\subsection{기관 및 시설 (Institutions and Facilities)}

\subsubsection{공공기관 (Government Institutions)}
\begin{itemize}
\setlength{\itemsep}{0pt}
\setlength{\parskip}{0pt}
\setlength{\parsep}{0pt}
\setlength{\topsep}{2pt}
\setlength{\partopsep}{0pt}
  \item 중앙행정·지방자치기관 (central and local government body)
  \item 국방기관 (defense agency)
  \item 치안및교정기관 (law enforcement and correctional institution)
  \item 소방기관 (fire protection agency)
  \item 금융기관 (financial agency)
\end{itemize}

\subsubsection{학술·교육기관 (Academic and Educational Institutions)}
\begin{itemize}
\setlength{\itemsep}{0pt}
\setlength{\parskip}{0pt}
\setlength{\parsep}{0pt}
\setlength{\topsep}{2pt}
\setlength{\partopsep}{0pt}
  \item 유아교육기관 (preschool institution)
  \item 학교 (school)
  \item 연구개발기관 (research and development institution)
\end{itemize}

\subsubsection{사회기반시설 (Social Infrastructure)}
\begin{itemize}
\setlength{\itemsep}{0pt}
\setlength{\parskip}{0pt}
\setlength{\parsep}{0pt}
\setlength{\topsep}{2pt}
\setlength{\partopsep}{0pt}
  \item 교통·안전시설 (transportation and safety facility)
  \item 에너지·자원시설 (energy and natural resource facility)
  \item 주민편의시설 (public convenience facility)
  \item 통신시설 (telecommunication infrastructure)
  \item 사회복지시설 (social welfare facility)
\end{itemize}

\subsubsection{의료시설 (Medical and Healthcare Facilities)}
\begin{itemize}
\setlength{\itemsep}{0pt}
\setlength{\parskip}{0pt}
\setlength{\parsep}{0pt}
\setlength{\topsep}{2pt}
\setlength{\partopsep}{0pt}
  \item 병·의원 (hospital and clinic)
  \item 공공의료시설 (public health facility)
  \item 회복지원시설 (rehabilitation support facility)
\end{itemize}

\subsubsection{사회·경제시설 (Social and Economic Facilities)}
\begin{itemize}
\setlength{\itemsep}{0pt}
\setlength{\parskip}{0pt}
\setlength{\parsep}{0pt}
\setlength{\topsep}{2pt}
\setlength{\partopsep}{0pt}
  \item 산업시설 (industrial complex)
  \item 주거시설 (housing facility)
  \item 종교시설 (place of worship)
\end{itemize}

\subsection{사업체 (Corporate Entities)}

\subsubsection{외식업 (Eating and Drinking Places)}
\begin{itemize}
\setlength{\itemsep}{0pt}
\setlength{\parskip}{0pt}
\setlength{\parsep}{0pt}
\setlength{\topsep}{2pt}
\setlength{\partopsep}{0pt}
  \item 식당 (restaurants)
  \item 주점 (bars and pubs)
  \item 카페 (cafe)
\end{itemize}

\subsubsection{도소매및유통업 (Retail and Distribution Industry)}
\begin{itemize}
\setlength{\itemsep}{0pt}
\setlength{\parskip}{0pt}
\setlength{\parsep}{0pt}
\setlength{\topsep}{2pt}
\setlength{\partopsep}{0pt}
  \item 식품판매업 (food retail business)
  \item 의약품판매업 (pharmaceutical retail business)
  \item 공산품판매업 (manufactured goods retail business)
  \item 동식물자원판매업 (animal and plant resource retail business)
  \item 종합물품판매업 (general merchandise retail business)
\end{itemize}

\subsubsection{운수·물류업 (Transportation and Logistics Industry)}
\begin{itemize}
\setlength{\itemsep}{0pt}
\setlength{\parskip}{0pt}
\setlength{\parsep}{0pt}
\setlength{\topsep}{2pt}
\setlength{\partopsep}{0pt}
  \item 화물운송·물류업 (freight transport and logistics business)
  \item 여객운송업 (passenger transport business)
\end{itemize}

\subsubsection{금융업 (Financial Industry)}
\begin{itemize}
\setlength{\itemsep}{0pt}
\setlength{\parskip}{0pt}
\setlength{\parsep}{0pt}
\setlength{\topsep}{2pt}
\setlength{\partopsep}{0pt}
  \item 대부업 (moneylending business)
  \item 보험업 (insurance business)
  \item 신탁업 (trust business)
  \item 증권업 (securities trading business)
  \item 여신금융업 (credit finance business)
  \item 은행업 (banking business)
\end{itemize}

\subsubsection{전문직 서비스업 (Professional Service Industry)}
\begin{itemize}
\setlength{\itemsep}{0pt}
\setlength{\parskip}{0pt}
\setlength{\parsep}{0pt}
\setlength{\topsep}{2pt}
\setlength{\partopsep}{0pt}
  \item 세무업 (tax service business)
  \item 회계업 (accounting business)
  \item 감정평가업 (appraisal and asset valuation business)
  \item 노무업 (labor affairs consultant business)
  \item 변호사업 (legal services business)
  \item 법무사업 (judicial scrivener business)
\end{itemize}

\subsubsection{건설·부동산업 (Construction and Real Estate Industry)}
\begin{itemize}
\setlength{\itemsep}{0pt}
\setlength{\parskip}{0pt}
\setlength{\parsep}{0pt}
\setlength{\topsep}{2pt}
\setlength{\partopsep}{0pt}
  \item 건설관리·설계업 (construction management and architectural design business)
  \item 건설시공업 (construction execution business)
  \item 부동산개발업 (real estate development business)
  \item 부동산중개업 (real estate brokerage business)
  \item 부동산임대·매매업 (real estate rental and sales business)
\end{itemize}

\subsubsection{정보통신업 (Information and Communications Industry)}
\begin{itemize}
\setlength{\itemsep}{0pt}
\setlength{\parskip}{0pt}
\setlength{\parsep}{0pt}
\setlength{\topsep}{2pt}
\setlength{\partopsep}{0pt}
  \item 언론관련산업 (media and press industry)
  \item 통신업 (telecommunications business)
  \item 출판업 (publishing business)
  \item 소프트웨어개발업 (software development business)
\end{itemize}

\subsubsection{제조업 (Manufacturing Industry)}
\begin{itemize}
\setlength{\itemsep}{0pt}
\setlength{\parskip}{0pt}
\setlength{\parsep}{0pt}
\setlength{\topsep}{2pt}
\setlength{\partopsep}{0pt}
  \item 식품제조업 (food manufacturing business)
  \item 의약품제조업 (pharmaceutical manufacturing business)
  \item 중간재제조업 (intermediate goods manufacturing business)
  \item 공산품제조업 (consumer goods manufacturing business)
  \item 중장비제조업 (heavy machinery manufacturing business)
\end{itemize}

\subsubsection{자원채취·생산업 (Resource Extraction and Production Industry)}
\begin{itemize}
\setlength{\itemsep}{0pt}
\setlength{\parskip}{0pt}
\setlength{\parsep}{0pt}
\setlength{\topsep}{2pt}
\setlength{\partopsep}{0pt}
  \item 농업 (agriculture)
  \item 축산업 (livestock farming)
  \item 수산업 (fishery)
  \item 광업 (mining)
  \item 임업 (forestry)
\end{itemize}

\subsubsection{숙박업 (Accommodation Industry)}
\begin{itemize}
\setlength{\itemsep}{0pt}
\setlength{\parskip}{0pt}
\setlength{\parsep}{0pt}
\setlength{\topsep}{2pt}
\setlength{\partopsep}{0pt}
  \item 대형숙박업 (hotel and resort business)
  \item 소형숙박업 (guesthouse and inn business)
\end{itemize}

\subsubsection{여가산업 (Leisure Industry)}
\begin{itemize}
\setlength{\itemsep}{0pt}
\setlength{\parskip}{0pt}
\setlength{\parsep}{0pt}
\setlength{\topsep}{2pt}
\setlength{\partopsep}{0pt}
  \item 오락·스포츠업 (entertainment and sports business)
  \item 문화·예술업 (culture and arts business)
\end{itemize}

\subsubsection{외모관리산업 (Beauty and Personal Care Industry)}
\begin{itemize}
\setlength{\itemsep}{0pt}
\setlength{\parskip}{0pt}
\setlength{\parsep}{0pt}
\setlength{\topsep}{2pt}
\setlength{\partopsep}{0pt}
  \item 두발피부미용업 (hair and skincare business)
  \item 체형관리업 (body care and management business)
  \item 목욕장업 (bathhouse business)
  \item 마사지업 (massage business)
\end{itemize}

\subsubsection{기타산업 (Other Service Industries)}
\begin{itemize}
\setlength{\itemsep}{0pt}
\setlength{\parskip}{0pt}
\setlength{\parsep}{0pt}
\setlength{\topsep}{2pt}
\setlength{\partopsep}{0pt}
  \item 교육서비스업 (education service business)
  \item 관혼상제서비스업 (ceremonial service business)
  \item 생활세척서비스업 (laundry and cleaning service)
  \item 동식물서비스업 (animal and plant care service business)
  \item 시설관리및임대서비스업 (facility management and rental service business)
  \item 수리서비스업 (repair service business)
  \item 기타서비스업 (other service business)
\end{itemize}

\subsection{조직 (Organizations)}

\subsubsection{사회·문화조직 (Social and Cultural Organization)}
\begin{itemize}
\setlength{\itemsep}{0pt}
\setlength{\parskip}{0pt}
\setlength{\parsep}{0pt}
\setlength{\topsep}{2pt}
\setlength{\partopsep}{0pt}
  \item 친목 (fellowship organization and social club)
  \item 가계 (family lineage organization)
  \item 종교 (religious organization)
  \item 사회복지 (social welfare organization)
\end{itemize}

\subsubsection{정치·경제조직 (Political and Economic Organization)}
\begin{itemize}
\setlength{\itemsep}{0pt}
\setlength{\parskip}{0pt}
\setlength{\parsep}{0pt}
\setlength{\topsep}{2pt}
\setlength{\partopsep}{0pt}
  \item 정치·의사결정 (political decision-making organization)
  \item 사회적 권익 (civil rights and advocacy organization)
  \item 경제적 권익 (economic rights and interest organization)
  \item 도시·주거 개발 (urban development and housing organization)
  \item 사회서비스 및 공익 (public service and community interest organization)
\end{itemize}

\subsubsection{국방조직 (Defense Organization)}
\begin{itemize}
\setlength{\itemsep}{0pt}
\setlength{\parskip}{0pt}
\setlength{\parsep}{0pt}
\setlength{\topsep}{2pt}
\setlength{\partopsep}{0pt}
  \item 군사편제 (military unit and formation)
\end{itemize}

\subsubsection{조직체계(Organizational Structure)}
\begin{itemize}
\setlength{\itemsep}{0pt}
\setlength{\parskip}{0pt}
\setlength{\parsep}{0pt}
\setlength{\topsep}{2pt}
\setlength{\partopsep}{0pt}
  \item 조직 내 세부부서(internal department and divisions)
  \item 조직내 업무·권한 등(internal duties and authorities)
\end{itemize}

\subsubsection{불법단체 (Illegal Organization)}
\begin{itemize}
\setlength{\itemsep}{0pt}
\setlength{\parskip}{0pt}
\setlength{\parsep}{0pt}
\setlength{\topsep}{2pt}
\setlength{\partopsep}{0pt}
  \item 형법상 범죄단체 (criminal organization under criminal law)
  \item 국보법상 범죄단체 (anti-state organization under the National Security Act)
\end{itemize}

\subsection{그외 사건관련장소 및 이벤트 (Other Incident-related Locations and Events)}

\subsubsection{시설부속장소 (Facility-affiliated Location)}
\begin{itemize}
\setlength{\itemsep}{0pt}
\setlength{\parskip}{0pt}
\setlength{\parsep}{0pt}
\setlength{\topsep}{2pt}
\setlength{\partopsep}{0pt}
  \item 시설부속공간 (ancillary space)
  \item 시설부속구조물 (on-site structures and installations)
\end{itemize}

\subsubsection{특정시설에 부속되지 않는 공간 (Unaffiliated Space)}
\begin{itemize}
\setlength{\itemsep}{0pt}
\setlength{\parskip}{0pt}
\setlength{\parsep}{0pt}
\setlength{\topsep}{2pt}
\setlength{\partopsep}{0pt}
  \item 건설현장 (construction cite)
  \item 선박 (vessel)
\end{itemize}

\subsubsection{이벤트 (Events)}
\begin{itemize}
\setlength{\itemsep}{0pt}
\setlength{\parskip}{0pt}
\setlength{\parsep}{0pt}
\setlength{\topsep}{2pt}
\setlength{\partopsep}{0pt}
  \item 공공 행사 (public event)
  \item 사업 및 과업 (project and assignment)
  \item 민간 행사 (private event)
\end{itemize}

\subsection{상품·서비스·콘텐츠 (Products, Services, and Contents)}

\subsubsection{일반상품 (Consumer Goods)}
\begin{itemize}
\setlength{\itemsep}{0pt}
\setlength{\parskip}{0pt}
\setlength{\parsep}{0pt}
\setlength{\topsep}{2pt}
\setlength{\partopsep}{0pt}
  \item 식품 (food products)
  \item 의약품 (pharmaceutical products)
  \item 반려동물 (pets and animals)
  \item 공산품 (manufactured products)
\end{itemize}

\subsubsection{금융상품(Financial Products)}
\begin{itemize}
\setlength{\itemsep}{0pt}
\setlength{\parskip}{0pt}
\setlength{\parsep}{0pt}
\setlength{\topsep}{2pt}
\setlength{\partopsep}{0pt}
  \item 투자상품 (investment products)
  \item 보험상품 (insurance products)
  \item 대출상품 (loan products)
\end{itemize}

\subsubsection{디지털상품 (Digital Products)}
\begin{itemize}
\setlength{\itemsep}{0pt}
\setlength{\parskip}{0pt}
\setlength{\parsep}{0pt}
\setlength{\topsep}{2pt}
\setlength{\partopsep}{0pt}
  \item 가상자산 (digital assets)
  \item 게임상품 (gaming items)
\end{itemize}

\subsubsection{디지털서비스 (Digital Services)}
\begin{itemize}
\setlength{\itemsep}{0pt}
\setlength{\parskip}{0pt}
\setlength{\parsep}{0pt}
\setlength{\topsep}{2pt}
\setlength{\partopsep}{0pt}
  \item 온라인커뮤니티 (online communities)
  \item 전자상거래 (e-commerce platforms)
  \item 기타사이트 (other websites)
  \item url (url)
  \item 소프트웨어 (software)
\end{itemize}

\subsubsection{사회문화콘텐츠 (Socio-Cultural Contents)}
\begin{itemize}
\setlength{\itemsep}{0pt}
\setlength{\parskip}{0pt}
\setlength{\parsep}{0pt}
\setlength{\topsep}{2pt}
\setlength{\partopsep}{0pt}
  \item 텍스트콘텐츠 (text contents)
  \item 이미지·영상콘텐츠 (image and video contents)
  \item 음원콘텐츠 (audio contents)
  \item 공연콘텐츠 (performance contents)
  \item 오락·스포츠콘텐츠 (entertainment and sports)
  \item 국가유산 (national heritage)
  \item 사건 (incidents)
  \item 과목 (academic subjects)
\end{itemize}

\subsubsection{지식재산 (Intellectual Property)}
\begin{itemize}
\setlength{\itemsep}{0pt}
\setlength{\parskip}{0pt}
\setlength{\parsep}{0pt}
\setlength{\topsep}{2pt}
\setlength{\partopsep}{0pt}
  \item 특허 (patents)
  \item 상표 (trademarks)
\end{itemize}

\section{Detailed Tokenization Scoring Function}
\label{app:tokenization_scoring}

Given a candidate tokenization sequence $X = (x_1, x_2, \dots, x_L)$, the proposed method selects the optimal sequence by maximizing a combined score. The score consists of two terms: an N-gram language model score and a boundary-aware score.

\begin{small}
\begin{equation*}
\operatorname{Score}(X) =
w_\mathrm{LM} \cdot \operatorname{Score}_\mathrm{LM}(X)
+
w_\mathrm{bound} \cdot \operatorname{Score}_\mathrm{bound}(X)
\end{equation*}
\end{small}

Here, $\operatorname{Score}_\mathrm{LM}$ is the language model score assigned by KenLM~\cite{heafield2011kenlm}. It measures how likely the candidate token sequence is based on N-gram statistics learned from the corpus. This term helps preserve tokenization patterns similar to those of the original tokenizer.

$\operatorname{Score}_\mathrm{bound}$ measures how well the candidate token sequence aligns with PII span boundaries. The weights $w_\mathrm{LM}$ and $w_\mathrm{bound}$ control the trade-off between preserving the original tokenizer behavior and encouraging boundary-aware tokenization.

The boundary-aware score is computed as the sum of token-level boundary scores:

\begin{small}
\begin{equation*}
\operatorname{Score}_\mathrm{bound}(X)=
\sum_{i=1}^{L}
b(x_i \mid x_{0:i-1})
\end{equation*}
\end{small}

where $b(x_i \mid x_{0:i-1})$ is the boundary score assigned to token $x_i$ given its preceding context. This score reflects how strongly the current token pattern is associated with PII span boundaries in the training corpus.

Similar to an N-gram language model, the boundary score is estimated using multiple context lengths with interpolation and backoff. For each token $x_i$, we consider contexts up to length $M$. In this work, we use up to trigram contexts.

\begin{small}
\begin{align*}
& b(x_i \mid x_{0:i-1}) \\
& \quad =
\sum_{j=1}^{M}
w_j
\cdot
\log
\left(
\frac{
C \cdot \operatorname{count}_\mathrm{bound}(x_i \mid x_{i-j:i-1})
}{
\operatorname{count}(x_i \mid x_{i-j:i-1})
}
+
1
\right)
\end{align*}
\end{small}

Here, $M$ is the maximum context length, and $w_j$ is the interpolation weight for context length $j$. The term $\operatorname{count}(x_i \mid x_{i-j:i-1})$ denotes how often the corresponding N-gram pattern appears in the corpus, while $\operatorname{count}_\mathrm{bound}(x_i \mid x_{i-j:i-1})$ counts how often the same pattern appears at PII span boundaries.

The constant $C$ controls the magnitude of the boundary signal. The logarithm prevents highly frequent boundary patterns from dominating the score, and the additive constant $+1$ ensures numerical stability.

Intuitively, an N-gram pattern receives a high boundary score when it frequently occurs near PII span boundaries relative to its overall frequency. Conversely, common token patterns that rarely correspond to de-identification boundaries receive lower boundary scores.

The final scoring function therefore favors tokenization candidates that satisfy both criteria: they remain plausible under the original tokenizer's distribution, while also producing token boundaries that are useful for precise de-identification.

In this paper, we use a 5-gram KenLM model for the language model component. During tokenization search, we retain the 256 highest-scoring candidate sequences at each step. We set the boundary scaling constant $C$ to 5 and the scoring weights $w_{\mathrm{LM}}$ and $w_{\mathrm{bound}}$ to 1.0 and 2.0, respectively.
\section{Metric Details}
\label{app:metric}

\paragraph{Counting TP, FP, FN, and TN.}
For a selected metric variant, TP, FP, FN, and TN are counted at the word-segment level. 
A gold PII word-segment is counted as a TP if it satisfies the selected coverage criterion and, for typed metrics, the predicted type matches the gold type. 
A gold PII word-segment is counted as an FN if it is not recovered or if it is recovered with the wrong type under typed evaluation. 
A non-PII word-segment with any PII prediction is counted as an FP, and a non-PII word-segment with no PII prediction is counted as a TN. 
Under typed evaluation, a wrong-type prediction on a gold PII word-segment is penalized as both an FN for the missed gold type and an FP for the incorrect predicted type. 
TN is not used in precision, recall, or F1-score:
\begin{equation}
\begin{aligned}
P &= \frac{\mathrm{TP}}{\mathrm{TP} + \mathrm{FP}}, \\ %\qquad
R &= \frac{\mathrm{TP}}{\mathrm{TP} + \mathrm{FN}}, \\
F1 &= \frac{2\,\mathrm{TP}}{2\,\mathrm{TP} + \mathrm{FP} + \mathrm{FN}}.
\end{aligned}
\end{equation}

% \paragraph{Aggregation across seeds.}
% When results are reported across multiple random seeds, we micro-aggregate by summing TP, FP, and FN across runs, then recomputing precision, recall, and F1-score from the aggregated counts.

\paragraph{Languages without whitespace delimiters.}
For languages without explicit whitespace delimiters, such as Chinese and Japanese, whitespace-based non-PII segments can become excessively long. 
In such cases, we recommend applying a morphological analyzer to non-PII regions and using morpheme-level segments. 
In Korean, whitespace-delimited units largely correspond to eojeols, and Sejong-based datasets report about 2 morphemes per eojeol~\cite{song2019korean}; we therefore use whitespace-delimited word-segments for reproducibility.
% \clearpage
\section{Model Configurations and Training Details}
\label{app:training}

% draft. 본문 다쓰고 정리

This appendix provides the architecture and training configurations for the \sys backbone models and the de-identification fine-tuning setup referenced in Section~\ref{sec:exp_setup}.

% \paragraph{Backbone pre-training.}
% All \sys models are pre-trained from scratch on a balanced bilingual Korean--English corpus with a 50K byte-level BPE vocabulary and a 4{,}096-token sequence length. The Korean and English portions are mixed with a balanced sampling strategy, and each model is trained on an equal split of Korean and English tokens. Table~\ref{tab:model_config} reports the per-model parameter counts, architecture sizes, and total pre-training token budgets, including the Korean/English split.

\paragraph{Backbone pre-training.}
All \sys models are pre-trained from scratch on a balanced bilingual Korean--English corpus with a 50K byte-level BPE vocabulary and a 4{,}096-token sequence length. The Korean portion is drawn from a Korean Common Crawl corpus filtered using the method proposed in \citep{park2025linelevelfilteringpretrainingcorpora}, while the English portion is drawn from FineWeb-Edu~\cite{penedo2024finewebdatasetsdecantingweb}. The two corpora are sampled in equal proportions, such that each model is trained on an equal number of Korean and English tokens. Table~\ref{tab:model_config} reports the per-model parameter counts, architecture sizes, and total pre-training token budgets, including the Korean/English split.

\paragraph{De-identification fine-tuning.}
For de-identification, we attach a linear token-classification head to each pretrained encoder and fine-tune with AdamW using bfloat16 precision and weight decay 0.01. Documents are converted into windows of 4{,}096 tokens with stride 128. The learning rate is $5\times10^{-5}$ for \textsc{150M} and \textsc{395M}, and $1\times10^{-5}$ for \textsc{750M} and \textsc{1B}. For each run, we select the checkpoint with the highest validation W. Strict Typed F1-score and report test scores from that checkpoint. The same optimization settings are used for the in-domain judgment and cross-domain general-document experiments, except that in-domain fine-tuning runs for 30 epochs and each cross-domain subset is trained for 3 epochs (Table~\ref{tab:model_config}).

\begin{table*}[!t]
\centering
\small
\setlength{\tabcolsep}{4pt}
\begin{tabular}{lrrrr}
\toprule
& \textsc{150M} & \textsc{395M} & \textsc{750M} & \textsc{1B} \\
\midrule
Layers & 22 & 28 & 32 & 22 \\
Hidden size & 768 & 1024 & 1152 & 1792 \\
\midrule
Vocabulary & \multicolumn{4}{c}{50K byte-level BPE} \\
Sequence length & \multicolumn{4}{c}{4,096} \\
Pre-training tokens & 10B (5/5) & 25B (12.5/12.5) & 40B (20/20) & 60B (30/30) \\
\midrule
Fine-tuning epochs (in-domain) & \multicolumn{4}{c}{30} \\
Fine-tuning epochs (cross-domain) & \multicolumn{4}{c}{3} \\
Optimizer & \multicolumn{4}{c}{AdamW, weight decay 0.01} \\
Precision & \multicolumn{4}{c}{bfloat16} \\
Window / stride & \multicolumn{4}{c}{4,096 / 128} \\
Effective batch size & 32 & 32 & 16 & 16 \\
Learning rate & $5\times10^{-5}$ & $5\times10^{-5}$ & $1\times10^{-5}$ & $1\times10^{-5}$ \\
\bottomrule
\end{tabular}%
\caption{Architecture and pre-training configurations of the \sys backbone models. Token budgets are reported in billions, with the Korean/English split in parentheses. Fine-tuning hyperparameters are shared across models except for the learning rate, which is $5\times10^{-5}$ for \textsc{150M}/\textsc{395M} and $1\times10^{-5}$ for \textsc{750M}/\textsc{1B}.}
\label{tab:model_config}
\end{table*}

\section{Replacement and Augmentation Details}
\label{app:replacement_augmentation}

\paragraph{Overview.}

The replacement pipeline converts inline-annotated Korean court judgments into surrogate-realized corpora and reconstructs character-level PII span annotations on the replaced text. The pipeline has three stages: (i) document-level annotation-boundary repair and label normalization, (ii) group-level type-compatible surrogate selection or generation, and (iii) context-aware Korean surface-form adjustment with character-level span reconstruction.

\paragraph{Entity inventory.}
We constructed a surrogate inventory aligned with the refined PII taxonomy. The inventory contains 1{,}800{,}316 lexicon-backed candidate entries and 68 rule-backed generator keys, grouped by top-level taxonomy category. Table~\ref{tab:surrogate-inventory-taxonomy-aligned} summarizes the inventory size.

\begin{table}[t]
\centering
\small
\setlength{\tabcolsep}{4pt}
\renewcommand{\arraystretch}{1.05}
\begin{tabular}{@{}lrr@{}}
\toprule
Taxonomy category & \makecell{Candidate\\entries} & \makecell{Generator\\rules} \\
\midrule
Direct identifiers & 67 & 13 \\
Incident-related numbers & 50 & 45 \\
Geographic information & 105{,}898 & 6 \\
Institutions and facilities & 279{,}542 & 0 \\
Corporate entities & 1{,}343{,}308 & 0 \\
Organizations & 38{,}775 & 1 \\
Other case-related loc./events & 7{,}533 & 0 \\
Products/services/contents & 25{,}143 & 3 \\
\midrule
Total & 1{,}800{,}316 & 68 \\
\bottomrule
\end{tabular}
\caption{Size of the surrogate inventory aligned with the top-level taxonomy in Figure~\ref{fig:pii-taxonomy-overview}. Category names are shortened for compactness. Candidate entries are counted from curated lexicon-backed resources, and generator rules correspond to normalized labels handled by rule-backed generators.}
\label{tab:surrogate-inventory-taxonomy-aligned}
\end{table}

Candidate sources were selected separately for each category. Business-related labels were populated from company and establishment name resources; institutional and facility labels from school, medical, public-institution, transportation, welfare, residential, and industrial-facility resources; geographic labels from administrative, address, road, and natural-feature resources; organization labels from political, economic, social, cultural, religious, military, and internal-department resources; and product, service, content, event, and place-descriptor labels from category-specific public directories and manually curated resources.

\paragraph{Candidate inventory curation.}
Candidate lists were compiled and cleaned before use. Entries were organized under the refined PII taxonomy, and ambiguous cases were handled through label-normalization rules, source-selection rules, and label-specific post-processing rules. Errors observed during dataset generation or model debugging were folded back into the candidate inventories, preprocessing rules, or post-processing rules.

\paragraph{Config-driven replacement.}
Each annotation is represented as $\lll$MENTION:LABEL[:SUB-LABEL]$\ggg$, where labels may be followed by colon-separated sub-labels. Before replacement, document-level preprocessing and config-level label preprocessing repair common annotation-boundary artifacts, drop labels that should not be used for replacement, and normalize labels through exact, wildcard, suffix-based, and compatibility rules. The normalized label is then mapped to a replacement rule.

\begin{itemize}
\item \textbf{Sampling from an inventory.} For labels with realistic external candidates, the system samples from the configured entity-inventory source. When multiple source files are available, stratified source-file sampling is used by default; concatenated sampling is used only when explicitly configured. Some broad or ambiguous labels also use nearby context to select a more specific source.
\item \textbf{Rule-based generation.} For structured values such as dates, ages, phone numbers, identifiers, transport numbers, and similar expressions, the system calls a seeded label-specific generator.
\item \textbf{Fallback or ignored cases.} Labels that should be preserved, skipped, or handled without an inventory candidate are controlled by explicit fallback rules. Depending on the configuration, missing candidates are ignored, treated as errors, or passed to a generator.
\end{itemize}

Replacement is reproducible given the replacement seed. The provider combines the seed with the document identifier and the annotation group key, so different documents and different seeds can receive different replacements. In the default grouping strategy, annotations are grouped by the mention--label key after preprocessing; repeated occurrences of the same mention with the same normalized label within a document therefore share one replacement. This is a surface-form consistency rule rather than a document-level coreference resolver.

\paragraph{Korean surface-form post-processing.}
Korean legal text often attaches particles, counters, suffixes, role markers, and type words directly to entity names. A replacement string that is correct in isolation can become unnatural or boundary-inconsistent after insertion into context. We therefore apply context-aware post-processors after surrogate selection or generation. These rules adjust or preserve surface markers such as institution types, school/facility suffixes, administrative-unit suffixes, station/road suffixes, phone-number wrappers, and other label-specific forms. The implementation also corrects following Korean particles such as \textit{i/ga}, \textit{eun/neun}, \textit{eul/reul}, \textit{gwa/wa}, and \textit{ro/euro}, and removes adjacent literal overlaps when a replacement already contains a suffix that appears immediately outside the annotation boundary. These steps keep the replaced text grammatical while aligning the gold span to the inserted replacement text.

\paragraph{Character-level dataset construction.}
The replacement output is rendered in plain-text mode for training. During reconstruction, the dataset builder records the begin and end character offsets of each usable inserted replacement in the replaced document text. If a generated replacement contains multiple inline annotations, the builder flattens the generated text and splits it into multiple character spans. Replacements that are ignored, missing, errored, or marked as not implemented are rendered as plain text but are not used as training spans.

The resulting records store both full replaced documents and line-centered context examples with neighboring context. Training labels are normalized after span construction. In the main setting, the training target uses the configured small-label level rather than the full replacement leaf label; replacement-only labels and configured aliases are handled by the training-target policy.

\paragraph{Single and per-epoch replacement.}
We used two replacement schedules. In the \textit{single} setting, a single replaced training corpus is generated from the training seed and reused for all epochs. In the \textit{per-epoch} setting, the training records are reloaded from an epoch-specific replacement dataset at each epoch. For the three training seeds used in our main experiments, the epoch-level replacement seeds follow a non-overlapping block schedule:

\[
s_{\mathrm{replace}}
= (s_{\mathrm{train}} - 1199) \times 10000 + 2000 + e.
\]

Here, $s_{\mathrm{train}}$ denotes the training seed, $e$ denotes the one-based epoch index, and $s_{\mathrm{replace}}$ denotes the replacement seed. Thus, train seeds 1200, 1201, and 1202 use epoch-level replacement seeds 12001--12030, 22001--22030, and 32001--32030, respectively. The validation and test records used for model selection and final evaluation are loaded from a fixed evaluation dataset, using a separate split seed, and remain unchanged across epochs and replacement seeds.

% KS: 다른곳에서 일부 해결됨 (appendix G 등)

% \section{Test set construction}
% \label{app:test_set_construction}

% \hl{===TODO===}: 세부 사항 확인하기. 참고해서 글 쓰기

% ========[train-test split]========

% - split 시 stratification 여부 : 사건 유형별 분포? PII label 분포?

% - train/test 비율 및 각 문서 수

% - train/test PII label 분포

% ========[general documents]========

% - 사용한 원천 corpus : CommonCrawl - 구체적으로 무엇?

% - 사용한 전처리 기법 : 찬우 논문 + ThunderLLM 때 기법 (단어 수 50~1,000,000 사이 / 평균 단어 길이 2~10 / 한글 글자 80\% 이상)

% - WebOrganizer/TopicClassifier-NoURL 기반 분류 -> format, subject, confidence 0.8 이상
% format: audio transcript / structured data / comment section / personal blog / user review / news || 
% subject: Crime \& Law / Social Life / Home \& Hobies / Education \& Jobs / Politics / Health

% - confidence가 여러 label에 대해 주어졌을 때: top-1 만 사용? multi-label? format, subject 모두 넘겨야? 못 넘는건 버렸는지? 주제는 왜 고려했는지?

% - annotator 에 의한 라벨링 (PII + 문서 종류[종류란게 무엇인지?]) -> 800개 : 정확한 숫자? 포함-배제 기준?

% - gpt5.4 로 1700개 생성 : 정확한 숫자? annotator 보고 few-shot 으로 생성한건지? 프롬프트 뭔지? 검수 여부? 생성때 주제 고려했는지?
\section{Surrogate Inventory Examples}
\label{app:surrogate_inventory}

This appendix complements Appendix~\ref{app:replacement_augmentation} by showing the lexicon-backed part of the surrogate inventory and representative surrogate surface forms.

The replacement pipeline uses a surrogate inventory aligned with the final PII taxonomy, with two resource classes. Lexicon-backed labels sample a same-type value from curated entity lists; rule-backed labels call deterministic generators for structured values such as names, dates, phone numbers, online identifiers, account numbers, registration numbers, and vehicle identifiers. Replacement is seeded, and repeated mention--label pairs receive the same surrogate within a document.

Table~\ref{tab:surrogate_inventory_examples} gives examples of surrogate surface forms. Structured identifiers, such as names, numbers, dates, and similar values, are synthetically generated and do not correspond to real individuals; lexicon-backed entries, such as institutions, businesses, organizations, and place names, are real entity names sampled from public resources. Some high-risk synthetic identifiers are shown partially masked, while the replacement corpus stores their complete values.

\begin{table*}[t]
\centering
\small
\setlength{\tabcolsep}{4pt}
\renewcommand{\arraystretch}{1.08}
\begin{tabular}{@{}p{0.24\linewidth}p{0.19\linewidth}p{0.39\linewidth}p{0.13\linewidth}@{}}
\toprule
Top-level PII category & Replacement source & Example surrogate surface forms & Note \\
\midrule
Direct identifiers
& Generator
& 윤건호 (Yun Geon-ho); \texttt{hyunwoo92@mail.test}; \texttt{198.51.100.177} (IP)
& Names and online identifiers \\

Incident-related numerical information
& Generator
& \texttt{4064-776-153368} (account no.); \texttt{870923-2******} (resident registration no.); \texttt{2000가합764234} (case no.)
& Structured identifiers \\

Geographic information
& Mixed generator and lexicon sample
& 세종대로 (Sejong-daero, road); \texttt{362-989번지} (lot no.); \texttt{11층} (11th floor)
& Address expressions \\

Institutions and facilities
& Lexicon sample
& 한국은행 (Bank of Korea); 금융감독원 (Financial Supervisory Service); 우체국 (post office)
& Public facilities \\

Corporate entities
& Lexicon sample
& 금산홍삼종합매장 (Geumsan red-ginseng store); 보은특산품가공식품매장 (Boeun local-food store); 롯데마트 아이오페 매장 (Lotte Mart IOPE counter)
& Businesses and stores \\

Organizations
& Lexicon sample
& 전국이통장연합회 (national village-head association); 금정구 새마을회 (Geumjeong Saemaul association); 노인생활과학연구소 (senior living-science institute)
& Civic and social groups \\

Other case-related locations and events
& Lexicon sample
& 공청회 (public hearing); 재개발정비사업 (redevelopment project); 연구과제 (research project)
& Events and projects \\

Products, services, and contents
& Lexicon sample
& 우리아파트론 (Woori apartment loan); 트라듀오서방정 (Traduo XR tablet); KB 한국 대표그룹주 증권 자투자신탁 (KB equity fund)
& Products and contents \\
\bottomrule
\end{tabular}
\caption{Examples of surrogate surface forms used by the replacement pipeline, organized by the top-level PII taxonomy. Structured identifiers are synthetically generated; lexicon-sampled entries are real public entity names. Korean values are glossed in parentheses.}
\label{tab:surrogate_inventory_examples}
\end{table*}
\section{General-Document Dataset}
\label{app:general_docs}
To evaluate cross-domain generalization, we construct a non-judgment Korean document benchmark containing PII.

\paragraph{Construction and provenance.}
The benchmark combines human-curated web documents with synthetic documents. The web-derived candidate pool was drawn from preprocessed Korean Common Crawl text. Preprocessing kept documents with at least 50 word-segments, an average word-segment length between 2.0 and 10.0, and a Korean-character ratio of at least 0.8. Candidate triage then used a WebOrganizer format classifier: we keep the top-1 predicted format label together with its softmax confidence and apply a format-confidence threshold of 0.8. Topic inference was not used as a keep/drop criterion. Human-curated documents were selected from this pool across news, online comments, online posts, and public-document sources, and were then manually reviewed; records with mixed snippets, label errors, severe quality problems, or duplicates were quarantined. Synthetic documents were added to cover PII types that are sparse in web text. All synthetic identifiers were produced by local rule-based format generators, so no real personal identifiers were inserted; the generator produced scenario-specific fragments, combined them into longer documents, and applied quality filters (e.g., invalid resident-registration-number lengths, implausible future dates, awkward particle attachments, and repeated or near-duplicate lines) followed by near-duplicate removal. After merging, document types were normalized through a content-based review pipeline and a final manual document-type review covering all documents. The resulting benchmark is split deterministically (seed 20260522) and stratified by category, subcategory, and source split; a small number of documents that failed span-level alignment during JSONL conversion were dropped, yielding the final 2{,}347-record benchmark described below.

\paragraph{Composition.}
The benchmark contains 2{,}347 documents, comprising 597 human-curated and 1{,}750 synthetic documents. Human-curated documents are drawn from non-judgment Korean web text; synthetic documents are added to cover practical document formats and PII configurations that are difficult to collect at scale from clean web text. The documents are split at the document level into train, validation, and test partitions of 1{,}211/136/1{,}000 documents, stratified by category, subcategory, and source type so that their proportions are preserved across splits. Table~\ref{tab:gendoc-category-source-split} reports the per-category, per-source breakdown across the three splits.

We organize documents along two complementary axes. The first is \emph{communicative purpose}, which groups documents into six categories: publications, administrative submissions, service requests, transactions, personal applications, and legal or business documents (Table~\ref{tab:gendoc-category-source-split}). Each category is further divided into functional subcategories that capture finer document intents---for example, publications split into articles, news, and notices/info, and personal applications split into education, identity/visa, and medical/insurance records; Table~\ref{tab:gendoc-subcategory-counts} lists all twelve subcategories with descriptions, and Table~\ref{tab:gendoc-subcategory-source-counts} reports their counts by source type. Human-curated documents are concentrated in publications, where suitable real-world examples are readily available; the remaining categories, which correspond to practical form-like documents that are hard to collect at scale, are populated almost entirely by synthetic documents. The second axis is the underlying \emph{document format}, recorded in document metadata; Table~\ref{tab:gendoc-format-source} reports the format-level distribution, which spans online posts, news, shopping and announcement pages, and other web document types.

\paragraph{Data-efficient adaptation subsets.}
For the experiments in Section~\ref{sec:exp_setup}, we evaluate models while increasing the number of training documents used for adaptation: a zero-shot setting with no general-document fine-tuning, followed by fine-tuning on nested training subsets of 25, 50, 100, 200, 500, and 1{,}000 documents (3 epochs each). The validation and test splits are kept fixed across all levels, and each subset is a stratified sample that preserves the category and source-type composition of the full training split, with smaller subsets contained in larger ones. Table~\ref{tab:gendoc-data-efficient-composition} reports their composition.

\paragraph{Annotation and label coverage.}
All documents are annotated with the same PII taxonomy used for the judgment corpus (Section~\ref{sec:Method:03_sub_PIItaxonomy}). The benchmark combines two sources: human-curated Korean web documents selected from Korean Common Crawl candidates containing PII, and synthetic documents generated to cover practical document formats and sparse identifier types without exposing private data. Because the benchmark is designed for non-judgment documents, its label distribution differs from that of court judgments. In particular, it covers 18 fine-grained PII types that can identify an individual on their own, such as personal names, contact information, online identifiers, and personal or account numbers. Context-dependent quasi-identifiers that are common in judgments, such as workplaces or incident locations, rarely appear in these documents and are not included. Table~\ref{tab:gendoc-pii-split-summary} summarizes the per-split document and span counts, and Table~\ref{tab:gendoc-pii-label-counts} reports span counts for each of the 18 labels. For evaluation, we restrict predictions to these 18 labels; predictions assigned to any label outside this set are treated as non-PII predictions, which prevents models from receiving credit for predicting judgment-specific labels that do not occur in this benchmark.

\begin{table*}[!t]
\centering
\scriptsize
\setlength{\tabcolsep}{3.5pt}
\resizebox{\textwidth}{!}{%
\begin{tabular}{lrrrrrrrrrrrr}
\toprule
 & \multicolumn{3}{c}{Train} & \multicolumn{3}{c}{Valid} & \multicolumn{3}{c}{Test} & \multicolumn{3}{c}{Total} \\
\cmidrule(lr){2-4} \cmidrule(lr){5-7} \cmidrule(lr){8-10} \cmidrule(lr){11-13}
Category & H & S & All & H & S & All & H & S & All & H & S & All \\
\midrule
Publications & 286 & 32 & 318 & 33 & 5 & 38 & 272 & 26 & 298 & 591 & 63 & 654 \\
Personal applications & 0 & 289 & 289 & 0 & 32 & 32 & 0 & 228 & 228 & 0 & 549 & 549 \\
Transactions & 0 & 230 & 230 & 0 & 25 & 25 & 0 & 180 & 180 & 0 & 435 & 435 \\
Service requests & 3 & 220 & 223 & 0 & 24 & 24 & 3 & 173 & 176 & 6 & 417 & 423 \\
Administrative submissions & 0 & 81 & 81 & 0 & 9 & 9 & 0 & 63 & 63 & 0 & 153 & 153 \\
Legal/business documents & 0 & 70 & 70 & 0 & 8 & 8 & 0 & 55 & 55 & 0 & 133 & 133 \\
\midrule
Total & 289 & 922 & 1{,}211 & 33 & 103 & 136 & 275 & 725 & 1{,}000 & 597 & 1{,}750 & 2{,}347 \\
\bottomrule
\end{tabular}%
}
\caption{Document categories and source types in the validated general-document split. H and S denote human-curated and synthetic documents.}
\label{tab:gendoc-category-source-split}
\end{table*}

\begin{table*}[!t]
\centering
\scriptsize
\setlength{\tabcolsep}{3.5pt}
\begin{tabular}{llrrrrrr}
\toprule
Category & Source & 25 & 50 & 100 & 200 & 500 & 1{,}000 \\
\midrule
Publications & Human & 6 & 13 & 25 & 49 & 119 & 238 \\
 & Synthetic & 0 & 0 & 1 & 3 & 12 & 24 \\
\addlinespace[1pt]
Personal applications & Human & 0 & 0 & 0 & 0 & 0 & 0 \\
 & Synthetic & 6 & 12 & 24 & 48 & 119 & 239 \\
\addlinespace[1pt]
Transactions & Human & 0 & 0 & 0 & 0 & 0 & 0 \\
 & Synthetic & 5 & 10 & 19 & 38 & 95 & 190 \\
\addlinespace[1pt]
Service requests & Human & 0 & 0 & 0 & 0 & 0 & 2 \\
 & Synthetic & 5 & 9 & 18 & 37 & 92 & 182 \\
\addlinespace[1pt]
Administrative submissions & Human & 0 & 0 & 0 & 0 & 0 & 0 \\
 & Synthetic & 2 & 3 & 7 & 13 & 34 & 67 \\
\addlinespace[1pt]
Legal/business documents & Human & 0 & 0 & 0 & 0 & 0 & 0 \\
 & Synthetic & 1 & 3 & 6 & 12 & 29 & 58 \\
\addlinespace[1pt]
\midrule
Total & Human & 6 & 13 & 25 & 49 & 119 & 240 \\
 & Synthetic & 19 & 37 & 75 & 151 & 381 & 760 \\
Total & All & 25 & 50 & 100 & 200 & 500 & 1{,}000 \\
\bottomrule
\end{tabular}
\caption{Document-category and source-type composition of the data-efficient adaptation training subsets. The subsets are nested stratified samples from the general-document training split; validation and test splits are fixed across all rows.}
\label{tab:gendoc-data-efficient-composition}
\end{table*}

\begin{table*}[!t]
\centering
\small
\setlength{\tabcolsep}{6pt}
\begin{tabular}{lrrr}
\toprule
Metadata document type & H & S & All \\
\midrule
Online post & 100 & 1{,}338 & 1{,}438 \\
Shopping info & 24 & 355 & 379 \\
News & 319 & 1 & 320 \\
Online comment & 77 & 17 & 94 \\
Announcement & 69 & 6 & 75 \\
Public notice & 2 & 33 & 35 \\
Interview & 2 & 0 & 2 \\
Report & 2 & 0 & 2 \\
Encyclopedia & 1 & 0 & 1 \\
Reservation info & 1 & 0 & 1 \\
\midrule
Total & 597 & 1{,}750 & 2{,}347 \\
\bottomrule
\end{tabular}
\caption{Format-level document type counts from the metadata of the validated general-document split. H and S denote human-curated and synthetic documents.}
\label{tab:gendoc-format-source}
\end{table*}

\begin{table*}[!t]
\centering
\small
\setlength{\tabcolsep}{3.5pt}
\begin{tabular}{lrrrr}
\toprule
Split & Documents & PII types & PII spans & Spans/doc \\
\midrule
Train & 1{,}211 & 18 & 11{,}034 & 9.11 \\
Validation & 136 & 18 & 1{,}203 & 8.85 \\
Test & 1{,}000 & 18 & 9{,}543 & 9.54 \\
\midrule
Total & 2{,}347 & 18 & 21{,}780 & 9.28 \\
\bottomrule
\end{tabular}
\caption{PII label coverage and span counts in the validated general-document split.}
\label{tab:gendoc-pii-split-summary}
\end{table*}

\begin{table*}[!t]
\centering
\scriptsize
\setlength{\tabcolsep}{1.5pt}
\renewcommand{\arraystretch}{1.02}
\begin{tabular*}{\columnwidth}{@{\extracolsep{\fill}}p{0.52\columnwidth}rrrr@{}}
\toprule
\textbf{PII label}
& \textbf{Train}
& \textbf{Validation}
& \textbf{Test}
& \textbf{Total} \\
\midrule
인명 (personal names)
& 3{,}182 & 333 & 3{,}187 & 6{,}702 \\
전화번호 (Phone number)
& 875 & 105 & 730 & 1{,}710 \\
업무관리번호 (task management number)
& 717 & 70 & 584 & 1{,}371 \\
이메일주소 (email address)
& 692 & 91 & 502 & 1{,}285 \\
닉네임 (Nickname)
& 653 & 67 & 547 & 1{,}267 \\
나이 (age)
& 593 & 58 & 506 & 1{,}157 \\
계좌번호 (bank account number)
& 574 & 62 & 446 & 1{,}082 \\
아이디 (ID)
& 566 & 72 & 419 & 1{,}057 \\
url (url)
& 512 & 68 & 453 & 1{,}033 \\
출생일자 (date of birth)
& 477 & 56 & 437 & 970 \\
결제수단관련번호 (payment method number)
& 447 & 50 & 375 & 872 \\
차량번호 (vehicle registration number)
& 317 & 31 & 257 & 605 \\
사업자등록번호 (business registration number)
& 307 & 36 & 222 & 565 \\
면허번호 (license number)
& 283 & 30 & 251 & 564 \\
주민등록번호 (resident registration number)
& 245 & 22 & 216 & 483 \\
법인등록번호 (corporate registration number)
& 264 & 28 & 155 & 447 \\
가상화폐개인지갑주소 (cryptocurrency personal wallet address)
& 188 & 16 & 152 & 356 \\
군번 (military service number)
& 142 & 8 & 104 & 254 \\
\midrule
\textbf{Total}
& \textbf{11{,}034}
& \textbf{1{,}203}
& \textbf{9{,}543}
& \textbf{21{,}780} \\
\bottomrule
\end{tabular*}
\caption{PII span counts by label in the validated general-document split.}
\label{tab:gendoc-pii-label-counts}
\end{table*}

\begin{table*}[!t]
\centering
\scriptsize
\setlength{\tabcolsep}{4pt}
\resizebox{\textwidth}{!}{%
\begin{tabular}{lllrrrr}
\toprule
Category & Subcategory & Description & Train & Validation & Test & Total \\
\midrule
Publications & Articles & Articles, reviews, essays, interviews, profiles, and reference-style text & 145 & 17 & 132 & 294 \\
Publications & News & News reports and press-release style reports & 103 & 12 & 107 & 222 \\
Publications & Notices/info & Notices, announcements, schedules, and public guidance & 70 & 9 & 59 & 138 \\
Administrative submissions & Submissions & Submissions, receipts, lists, surveys, and administrative forms & 81 & 9 & 63 & 153 \\
Service requests & Repair/support & Repair, installation, failure reports, and customer support records & 94 & 10 & 74 & 178 \\
Service requests & Reservations & Reservations, facility use, rentals, visits, and schedule changes & 129 & 14 & 102 & 245 \\
Transactions & Payments & Payment, refund, account, card, deposit, and receipt records & 156 & 17 & 122 & 295 \\
Transactions & Trade/delivery & Orders, delivery, purchase, sales, and trade dispute records & 74 & 8 & 58 & 140 \\
Personal applications & Education & School, academy, scholarship, course, and care-related records & 105 & 12 & 83 & 200 \\
Personal applications & Identity/visa & Identity, passport, visa, residence, welfare, and qualification records & 119 & 13 & 94 & 226 \\
Personal applications & Medical/insurance & Medical, hospital, insurance claim, license, and certificate records & 65 & 7 & 51 & 123 \\
Legal/business documents & Legal/business & Registration, legal notice, contract, business, and real-estate records & 70 & 8 & 55 & 133 \\
\midrule
Total & & & 1{,}211 & 136 & 1{,}000 & 2{,}347 \\
\bottomrule
\end{tabular}%
}
\caption{Functional subcategories in the validated general-document split.}
\label{tab:gendoc-subcategory-counts}
\end{table*}

\begin{table*}[!t]
\centering
\scriptsize
\setlength{\tabcolsep}{3.5pt}
\resizebox{\textwidth}{!}{%
\begin{tabular}{llrrrrrrrrr}
\toprule
 & & \multicolumn{3}{c}{Train} & \multicolumn{3}{c}{Validation} & \multicolumn{3}{c}{Test} \\
\cmidrule(lr){3-5} \cmidrule(lr){6-8} \cmidrule(lr){9-11}
Category & Subcategory & H & S & All & H & S & All & H & S & All \\
\midrule
Publications & Articles & 141 & 4 & 145 & 16 & 1 & 17 & 129 & 3 & 132 \\
Publications & News & 100 & 3 & 103 & 11 & 1 & 12 & 104 & 3 & 107 \\
Publications & Notices/info & 45 & 25 & 70 & 6 & 3 & 9 & 39 & 20 & 59 \\
Administrative submissions & Submissions & 0 & 81 & 81 & 0 & 9 & 9 & 0 & 63 & 63 \\
Service requests & Repair/support & 3 & 91 & 94 & 0 & 10 & 10 & 2 & 72 & 74 \\
Service requests & Reservations & 0 & 129 & 129 & 0 & 14 & 14 & 1 & 101 & 102 \\
Transactions & Payments & 0 & 156 & 156 & 0 & 17 & 17 & 0 & 122 & 122 \\
Transactions & Trade/delivery & 0 & 74 & 74 & 0 & 8 & 8 & 0 & 58 & 58 \\
Personal applications & Education & 0 & 105 & 105 & 0 & 12 & 12 & 0 & 83 & 83 \\
Personal applications & Identity/visa & 0 & 119 & 119 & 0 & 13 & 13 & 0 & 94 & 94 \\
Personal applications & Medical/insurance & 0 & 65 & 65 & 0 & 7 & 7 & 0 & 51 & 51 \\
Legal/business documents & Legal/business & 0 & 70 & 70 & 0 & 8 & 8 & 0 & 55 & 55 \\
\bottomrule
\end{tabular}%
}
\caption{Functional subcategory counts by source type. H and S denote human-curated and synthetic documents.}
\label{tab:gendoc-subcategory-source-counts}
\end{table*}

\section{Full Experimental Results}
\label{app:full_results}

This appendix reports the full in-domain and cross-domain results that the main paper summarizes for compactness.

\paragraph{In-domain, both replacement schedules.}
Table~\ref{tab:full-indomain-results} reports the complete in-domain test results on Korean court judgments, including the \textit{single}-replacement schedule omitted from the main text (Table~\ref{tab:main-indomain-results}). Across all model sizes and both tokenizers, per-epoch replacement is consistently, if modestly, better than single replacement, since fresh surrogate surface forms each epoch reduce overfitting to any one realization; we therefore adopt per-epoch replacement as the default. As in the main results, the proposed tokenizer outperforms BPE under both schedules and at every model size.

\begin{table*}[!tp]
\centering
\begingroup
\small
\setlength{\tabcolsep}{4pt}
\renewcommand{\arraystretch}{1.12}
\resizebox{\textwidth}{!}{%
\begin{tabular}{l l l rrr rrr rrr rrr}
\toprule
\multirow{2}{*}{Model} & \multirow{2}{*}{Tokenizer} & \multirow{2}{*}{Replacement}
& \multicolumn{3}{c}{W. Strict Typed}
& \multicolumn{3}{c}{W. Partial Typed}
& \multicolumn{3}{c}{W. Strict Binary}
& \multicolumn{3}{c}{W. Partial Binary} \\
\cmidrule(lr){4-6}\cmidrule(lr){7-9}\cmidrule(lr){10-12}\cmidrule(lr){13-15}
& & & F1 & P & R & F1 & P & R & F1 & P & R & F1 & P & R \\
\midrule
T-370M & Mecab-ko+BPE & --
& n/a & n/a & n/a & n/a & n/a & n/a
& 0.613 & 0.545 & 0.700 & 0.688 & 0.588 & 0.831 \\
T-800M & Mecab-ko+BPE & --
& n/a & n/a & n/a & n/a & n/a & n/a
& 0.697 & 0.601 & 0.831 & 0.738 & 0.621 & 0.908 \\
T-1.5B & Mecab-ko+BPE & --
& n/a & n/a & n/a & n/a & n/a & n/a
& 0.720 & 0.630 & 0.839 & 0.761 & 0.650 & 0.916 \\
\addlinespace[2pt]\midrule
\multirow{4}{*}{150M}
 & BPE           & single    & 0.744 & 0.663 & 0.846 & 0.770 & 0.676 & 0.894 & 0.775 & 0.694 & 0.877 & 0.798 & 0.704 & 0.921 \\
 & BPE           & per-epoch & 0.760 & 0.661 & 0.894 & 0.779 & 0.670 & 0.932 & 0.785 & 0.685 & 0.920 & 0.802 & 0.692 & 0.953 \\
 & \sys tok. & single    & 0.853 & 0.856 & 0.851 & 0.879 & 0.862 & 0.897 & 0.886 & 0.895 & 0.878 & 0.911 & 0.900 & 0.923 \\
 & \sys tok. & per-epoch & 0.867 & 0.851 & 0.884 & 0.886 & 0.856 & 0.919 & 0.890 & 0.877 & 0.904 & 0.908 & 0.881 & 0.937 \\
\addlinespace[2pt]
\multirow{4}{*}{395M}
 & BPE           & single    & 0.749 & 0.670 & 0.848 & 0.773 & 0.681 & 0.894 & 0.776 & 0.697 & 0.875 & 0.798 & 0.707 & 0.916 \\
 & BPE           & per-epoch & 0.761 & 0.669 & 0.883 & 0.781 & 0.678 & 0.920 & 0.783 & 0.689 & 0.905 & 0.799 & 0.697 & 0.937 \\
 & \sys tok. & single    & 0.852 & 0.853 & 0.851 & 0.877 & 0.859 & 0.896 & 0.882 & 0.887 & 0.876 & 0.903 & 0.892 & 0.915 \\
 & \sys tok. & per-epoch & 0.876 & 0.870 & 0.881 & 0.897 & 0.875 & 0.921 & 0.897 & 0.894 & 0.899 & 0.917 & 0.898 & 0.936 \\
\addlinespace[2pt]
\multirow{4}{*}{750M}
 & BPE           & single    & 0.762 & 0.683 & 0.863 & 0.788 & 0.694 & 0.910 & 0.787 & 0.707 & 0.887 & 0.809 & 0.716 & 0.930 \\
 & BPE           & per-epoch & 0.770 & 0.675 & 0.895 & 0.790 & 0.684 & 0.935 & 0.792 & 0.696 & 0.918 & 0.809 & 0.704 & 0.952 \\
 & \sys tok. & single    & 0.867 & 0.867 & 0.867 & 0.892 & 0.872 & 0.912 & 0.895 & 0.901 & 0.890 & 0.919 & 0.905 & 0.933 \\
 & \sys tok. & per-epoch & 0.877 & 0.864 & 0.891 & 0.900 & 0.870 & 0.932 & 0.901 & 0.892 & 0.911 & 0.922 & 0.896 & 0.949 \\
\addlinespace[2pt]
\multirow{4}{*}{1B}
 & BPE           & single    & 0.762 & 0.677 & 0.871 & 0.785 & 0.688 & 0.915 & 0.786 & 0.701 & 0.895 & 0.807 & 0.710 & 0.935 \\
 & BPE           & per-epoch & 0.771 & 0.671 & 0.906 & 0.790 & 0.680 & 0.942 & 0.791 & 0.689 & 0.927 & 0.806 & 0.696 & 0.957 \\
 & \sys tok. & single    & 0.868 & 0.864 & 0.873 & 0.891 & 0.869 & 0.914 & 0.896 & 0.896 & 0.896 & 0.916 & 0.900 & 0.934 \\
 & \sys tok. & per-epoch & 0.878 & 0.856 & 0.901 & 0.898 & 0.861 & 0.938 & 0.902 & 0.882 & 0.923 & 0.918 & 0.885 & 0.954 \\
\bottomrule
\end{tabular}%
}
\endgroup
\caption{Full in-domain test results on Korean court judgments under both single and per-epoch replacement. Scores are seed-micro aggregates over seeds 1200, 1201, and 1202. Strict requires full non-space coverage of a gold PII word-segment; Partial requires any non-space coverage. Typed additionally requires the predicted PII type to match the gold type; Binary ignores type. T-370M/800M/1.5B denote released Thunder-DeID checkpoints; their label inventory differs from ours, so typed metrics are not defined (n/a).}
\label{tab:full-indomain-results}
\end{table*}

\paragraph{Thunder-DeID under matched training data.}
To further control for differences in the training corpus, we retrain
Thunder-DeID~\cite{thunder-deid} on the same 4,619-document \sys
training subset used in the matched analysis in
Table~\ref{tab:taxonomy_matched}. For each Thunder-DeID model, we
initialize from the corresponding pretrained checkpoint and retain its
Mecab-ko+BPE tokenizer and training pipeline. Both Thunder-DeID and
\sys are evaluated on the same 574-document judgment test set using
the \sys annotation inventory, allowing both W.\ Strict Typed and
W.\ Strict Binary F1 to be compared directly.

\begin{table}[t]
\centering
\footnotesize
\setlength{\tabcolsep}{3.2pt}
\renewcommand{\arraystretch}{1.06}
\begin{tabular}{@{}llcc@{}}
\toprule
\textbf{Metric}
& \textbf{Model sizes}
& \shortstack{\textbf{Thunder-}\\\textbf{DeID}}
& \shortstack{\textbf{Thunder-}\\\textbf{DeID+}} \\
\midrule
\multirow{3}{*}{\shortstack[l]{W. Strict\\Typed F1}}
& 370M / 150M & 0.805 & \textbf{0.855} \\
& 800M / 395M & 0.810 & \textbf{0.867} \\
& 1.5B / 1B   & 0.795 & \textbf{0.877} \\
\midrule
\multirow{3}{*}{\shortstack[l]{W. Strict\\Binary F1}}
& 370M / 150M & 0.841 & \textbf{0.887} \\
& 800M / 395M & 0.852 & \textbf{0.891} \\
& 1.5B / 1B   & 0.840 & \textbf{0.906} \\
\bottomrule
\end{tabular}
\caption{Comparison with Thunder-DeID after fine-tuning on the same
4,619-document Thunder-DeID+ training subset. Both systems are evaluated
on the same 574-document judgment test set using the Thunder-DeID+
annotation inventory. Model-size pairs are ordered as
Thunder-DeID / Thunder-DeID+. The two model families use different
backbones and are not parameter-matched.}
\label{tab:thunder-matched-training}
\end{table}

Fine-tuning Thunder-DeID on the \sys training data substantially
improves the baseline, indicating that differences in the original
training corpora account for part of the performance gap. Nevertheless,
\sys remains consistently stronger across all three size tiers,
with gains of 0.050--0.082 W.\ Strict Typed F1 and
0.039--0.066 W.\ Strict Binary F1. Because the model architectures,
tokenizers, and parameter counts are not identical, this experiment
should be interpreted as a matched-training-data system comparison
rather than an architecture-only ablation.

\paragraph{Cross-domain, full adaptation curve.}
Tables~\ref{tab:gendoc-data-efficient-f1pr-results} and~\ref{tab:gendoc-data-efficient-strict-binary-with-thunder} report the full data-efficient cross-domain adaptation curves on the general-document test split across all subset sizes $\{0, 25, 50, 100, 200, 500, 1{,}000\}$. Table~\ref{tab:gendoc-data-efficient-f1pr-results} reports W.\ Strict Typed F1/P/R, corresponding to the primary metric summarized in the main paper at representative subset sizes (Table~\ref{tab:gendoc-typed-compact}). Table~\ref{tab:gendoc-data-efficient-strict-binary-with-thunder} reports W.\ Strict Binary F1/P/R over the same subset grid, with the released Thunder-DeID checkpoints (T-370M, T-800M, and T-1.5B) included as zero-shot references under the binary metric. Because the Thunder-DeID label inventory differs from ours, only binary metrics are reported for those checkpoints, and they appear only at column~$0$. All scores are seed-micro aggregates over seeds 1200, 1201, and 1202; our rows use judgment-trained per-epoch checkpoints, with validation and test fixed across all subset sizes.

\begin{table*}[!tp]
\centering
\tiny
\setlength{\tabcolsep}{2.0pt}
\renewcommand{\arraystretch}{1.05}
\resizebox{\textwidth}{!}{%
\begin{tabular}{llrrrrrrrrrrrrrrrrrrrrr}
\toprule
\multirow{2}{*}{Model} & \multirow{2}{*}{Tokenizer} & \multicolumn{3}{c}{0} & \multicolumn{3}{c}{25} & \multicolumn{3}{c}{50} & \multicolumn{3}{c}{100} & \multicolumn{3}{c}{200} & \multicolumn{3}{c}{500} & \multicolumn{3}{c}{1{,}000} \\
\cmidrule(lr){3-5} \cmidrule(lr){6-8} \cmidrule(lr){9-11} \cmidrule(lr){12-14} \cmidrule(lr){15-17} \cmidrule(lr){18-20} \cmidrule(lr){21-23}
& & F1 & P & R & F1 & P & R & F1 & P & R & F1 & P & R & F1 & P & R & F1 & P & R & F1 & P & R \\
\midrule
\multirow{2}{*}{150M} & BPE & 0.407 & 0.346 & 0.492 & 0.557 & 0.479 & 0.664 & 0.647 & 0.621 & 0.675 & 0.689 & 0.720 & 0.661 & 0.761 & 0.757 & 0.764 & 0.856 & 0.840 & 0.872 & 0.876 & 0.861 & 0.891 \\
 & \sys tok. & 0.453 & 0.402 & 0.519 & 0.597 & 0.530 & 0.684 & 0.679 & 0.677 & 0.681 & 0.727 & 0.769 & 0.689 & 0.790 & 0.807 & 0.775 & 0.876 & 0.897 & 0.857 & 0.894 & 0.898 & 0.890 \\
\addlinespace[2pt]
\multirow{2}{*}{395M} & BPE & 0.448 & 0.398 & 0.511 & 0.666 & 0.627 & 0.711 & 0.736 & 0.741 & 0.731 & 0.809 & 0.801 & 0.818 & 0.859 & 0.857 & 0.860 & 0.876 & 0.868 & 0.883 & 0.888 & 0.881 & 0.894 \\
 & \sys tok. & 0.491 & 0.464 & 0.520 & 0.676 & 0.624 & 0.737 & 0.742 & 0.770 & 0.716 & 0.818 & 0.834 & 0.803 & 0.879 & 0.901 & 0.858 & 0.896 & 0.910 & 0.883 & 0.907 & 0.923 & 0.892 \\
\addlinespace[2pt]
\multirow{2}{*}{750M} & BPE & 0.451 & 0.401 & 0.516 & 0.727 & 0.715 & 0.740 & 0.787 & 0.772 & 0.803 & 0.869 & 0.875 & 0.863 & 0.888 & 0.876 & 0.899 & 0.901 & 0.878 & 0.925 & 0.907 & 0.893 & 0.921 \\
 & \sys tok. & 0.485 & 0.448 & 0.529 & 0.743 & 0.737 & 0.749 & 0.814 & 0.814 & 0.814 & 0.889 & 0.908 & 0.871 & 0.908 & 0.917 & 0.899 & 0.924 & 0.922 & 0.926 & 0.930 & 0.935 & 0.924 \\
\addlinespace[2pt]
\multirow{2}{*}{1B} & BPE & 0.462 & 0.395 & 0.554 & 0.705 & 0.677 & 0.736 & 0.803 & 0.786 & 0.821 & 0.877 & 0.865 & 0.890 & 0.892 & 0.887 & 0.897 & 0.905 & 0.884 & 0.928 & 0.911 & 0.897 & 0.927 \\
 & \sys tok. & 0.494 & 0.442 & 0.560 & 0.744 & 0.762 & 0.728 & 0.814 & 0.824 & 0.805 & 0.891 & 0.890 & 0.893 & 0.909 & 0.920 & 0.898 & 0.925 & 0.921 & 0.929 & 0.933 & 0.937 & 0.928 \\
\bottomrule
\end{tabular}%
}
\caption{Full data-efficient cross-domain adaptation on the general-document test split. Each judgment-trained model uses per-epoch replacement and is adapted on nested stratified subsets of the general-document training split; column $0$ is zero-shot. Values are W.\ Strict Typed F1, precision (P), and recall (R), computed as seed-micro aggregates over seeds 1200/1201/1202. This is the full-coverage version of Table~\ref{tab:gendoc-typed-compact}.}
\label{tab:gendoc-data-efficient-f1pr-results}
\end{table*}

\begin{table*}[!tp]
\centering
\tiny
\setlength{\tabcolsep}{1.9pt}
\renewcommand{\arraystretch}{1.02}
\resizebox{\textwidth}{!}{%
\begin{tabular}{llrrrrrrrrrrrrrrrrrrrrr}
\toprule
\multirow{2}{*}{Model} & \multirow{2}{*}{Tokenizer}
& \multicolumn{3}{c}{0}
& \multicolumn{3}{c}{25}
& \multicolumn{3}{c}{50}
& \multicolumn{3}{c}{100}
& \multicolumn{3}{c}{200}
& \multicolumn{3}{c}{500}
& \multicolumn{3}{c}{1{,}000} \\
\cmidrule(lr){3-5} \cmidrule(lr){6-8} \cmidrule(lr){9-11} \cmidrule(lr){12-14} \cmidrule(lr){15-17} \cmidrule(lr){18-20} \cmidrule(lr){21-23}
& & F1 & P & R & F1 & P & R & F1 & P & R & F1 & P & R & F1 & P & R & F1 & P & R & F1 & P & R \\
\midrule
T-370M & Mecab-ko+BPE
& 0.226 & 0.228 & 0.223
& -- & -- & --
& -- & -- & --
& -- & -- & --
& -- & -- & --
& -- & -- & --
& -- & -- & -- \\
T-800M & Mecab-ko+BPE
& 0.484 & 0.427 & 0.559
& -- & -- & --
& -- & -- & --
& -- & -- & --
& -- & -- & --
& -- & -- & --
& -- & -- & -- \\
T-1.5B & Mecab-ko+BPE
& 0.507 & 0.492 & 0.523
& -- & -- & --
& -- & -- & --
& -- & -- & --
& -- & -- & --
& -- & -- & --
& -- & -- & -- \\
\midrule
150M & BPE
& 0.576 & 0.500 & 0.679
& 0.693 & 0.602 & 0.816
& 0.768 & 0.744 & 0.793
& 0.808 & 0.850 & 0.770
& 0.841 & 0.842 & 0.840
& 0.871 & 0.857 & 0.885
& 0.886 & 0.873 & 0.900 \\
150M & \sys tok.
& 0.602 & 0.544 & 0.674
& 0.722 & 0.647 & 0.817
& 0.794 & 0.800 & 0.789
& 0.834 & 0.888 & 0.787
& 0.864 & 0.886 & 0.843
& 0.890 & 0.913 & 0.868
& 0.905 & 0.911 & 0.899 \\
\addlinespace[2pt]
395M & BPE
& 0.593 & 0.539 & 0.660
& 0.756 & 0.717 & 0.800
& 0.807 & 0.816 & 0.797
& 0.847 & 0.841 & 0.852
& 0.869 & 0.870 & 0.869
& 0.884 & 0.878 & 0.890
& 0.896 & 0.891 & 0.902 \\
395M & \sys tok.
& 0.628 & 0.609 & 0.648
& 0.764 & 0.710 & 0.826
& 0.826 & 0.860 & 0.794
& 0.865 & 0.886 & 0.845
& 0.890 & 0.914 & 0.867
& 0.904 & 0.919 & 0.890
& 0.915 & 0.933 & 0.899 \\
\addlinespace[2pt]
750M & BPE
& 0.610 & 0.550 & 0.685
& 0.815 & 0.806 & 0.824
& 0.855 & 0.842 & 0.870
& 0.884 & 0.892 & 0.877
& 0.895 & 0.885 & 0.905
& 0.909 & 0.888 & 0.931
& 0.914 & 0.902 & 0.927 \\
750M & \sys tok.
& 0.632 & 0.593 & 0.677
& 0.849 & 0.847 & 0.850
& 0.881 & 0.883 & 0.878
& 0.904 & 0.925 & 0.883
& 0.916 & 0.927 & 0.906
& 0.931 & 0.930 & 0.932
& 0.938 & 0.944 & 0.931 \\
\addlinespace[2pt]
1B & BPE
& 0.611 & 0.530 & 0.719
& 0.815 & 0.787 & 0.845
& 0.872 & 0.856 & 0.889
& 0.891 & 0.881 & 0.902
& 0.900 & 0.896 & 0.903
& 0.914 & 0.894 & 0.935
& 0.919 & 0.905 & 0.933 \\
1B & \sys tok.
& 0.617 & 0.560 & 0.688
& 0.838 & 0.865 & 0.813
& 0.884 & 0.897 & 0.871
& 0.909 & 0.909 & 0.909
& 0.918 & 0.930 & 0.906
& 0.932 & 0.929 & 0.935
& 0.939 & 0.945 & 0.934 \\
\bottomrule
\end{tabular}%
}
\caption{Full W.\ Strict Binary F1-score/P/R results on the general-document test split. T-370M/800M/1.5B denote released Thunder-DeID checkpoints, included as zero-shot detection-level references because their label inventory differs from ours. Our rows use judgment-trained per-epoch checkpoints adapted on nested stratified subsets of the general-document training split; column $0$ is zero-shot.}
\label{tab:gendoc-data-efficient-strict-binary-with-thunder}
\end{table*}

\paragraph{Human-curated and synthetic test subsets.}
The general-document test set contains 275 human-curated and 725 synthetic documents. We therefore report results separately by source. Across model sizes, W. Strict Typed F1 ranges from 0.739 to 0.836 on the human-curated subset and from 0.997 to 0.999 on the synthetic subset. The synthetic documents mainly cover structured identifiers and form-like document types that are difficult to collect safely at scale, whereas the human-curated subset consists of naturally occurring Korean web text. The results on the human-curated subset show that the model also transfers to naturally occurring non-judgment documents.

\begin{table}[t]
\centering
\footnotesize
\setlength{\tabcolsep}{3.3pt}
\renewcommand{\arraystretch}{1.05}
\begin{tabular*}{\columnwidth}{@{\extracolsep{\fill}}lrrrr@{}}
\toprule
\multirow{2}{*}{\textbf{PII type}}
& \multicolumn{2}{c}{\textbf{Human-curated}}
& \multicolumn{2}{c}{\textbf{Synthetic}} \\
\cmidrule(lr){2-3}
\cmidrule(lr){4-5}
& \textbf{Gold} & \textbf{F1}
& \textbf{Gold} & \textbf{F1} \\
\midrule
Email address
& 95  & 0.990 & 407 & 1.000 \\
Phone number
& 64  & 0.916 & 666 & 0.999 \\
URL
& 112 & 0.910 & 341 & 1.000 \\
Account number
& 4   & 0.778 & 442 & 1.000 \\
Age
& 165 & 0.701 & 341 & 0.999 \\
Date of birth
& 35  & 0.579 & 402 & 1.000 \\
ID
& 28  & 0.543 & 391 & 0.999 \\
Nickname
& 244 & 0.412 & 303 & 0.999 \\
Business registration number
& 1   & 1.000 & 221 & 1.000 \\
\bottomrule
\end{tabular*}
\caption{Per-type W. Strict Typed F1 of the 1B \sys model on PII types appearing in both the human-curated and synthetic subsets of the general-document test set. Gold denotes the number of gold PII spans in each subset. Personal names are omitted because the synthetic test subset contains no personal-name instances. Results for account numbers and business registration numbers on the human-curated subset are based on only four and one gold instances, respectively.}
\label{tab:general-source-labelwise}
\end{table}

As shown in Table~\ref{tab:general-source-labelwise}, the difference between the two sources is also present at the PII-type level. On the synthetic subset, F1 is 0.999--1.000 for every shared type. On the human-curated subset, email addresses, phone numbers, and URLs remain above 0.90 F1, while age, dates of birth, IDs, and nicknames are more difficult.

\clearpage
\twocolumn

% 페이지 오류 수정중
\section{Tokenization Overhead}
\label{app:runtime_overhead}

To assess the computational cost of improving clean PII-span coverage, we measure the runtime overhead of the proposed tokenization method. Table\mbox{~\ref{tab:tokenization_runtime}} reports the mean tokenization latency per document and the character-level throughput of each tokenizer.

\begin{table}[h]
\centering
\small
\renewcommand{\arraystretch}{1.05}
\begin{tabular*}{\columnwidth}{@{\extracolsep{\fill}}lrr@{}}
\toprule
\textbf{Tokenizer}
& \shortstack{\textbf{Mean latency}\\\textbf{(ms/doc)} $\downarrow$}
& \shortstack{\textbf{Throughput}\\\textbf{(chars/s)} $\uparrow$} \\
\midrule
BPE           & 3.27  & 1,379,217 \\
MeCab+BPE     & 45.58 & 99,243    \\
\textbf{\sys} & \textbf{21.22} & \textbf{213,127} \\
\bottomrule
\end{tabular*}
\caption{Runtime comparison of the tokenization methods. Mean latency denotes the average tokenization time per document, and throughput denotes the number of input characters processed per second.}
\label{tab:tokenization_runtime}
\end{table}

Because \sys uses an $N$-gram language model to score candidate tokenizations during the search process, it incurs greater computational overhead than standard BPE, requiring 21.22 ms per document compared with 3.27 ms for BPE. Nevertheless, \sys requires less than half the processing time of MeCab+BPE and achieves approximately 2.15 times higher throughput, processing 213,127 characters per second compared with 99,243 characters per second. Given that MeCab-based tokenization is commonly used for large-scale Korean document processing, these results suggest that the additional boundary-aware computation introduced by \sys is not a critical bottleneck and that the proposed tokenizer remains practical for large-scale document processing.
\section{Per-type Error Analysis}
\label{app:per-type-analysis}

We report per-type W. Strict Typed precision, recall, and F1 to examine how performance varies across PII types. Scores are micro-aggregated across three seeds, while train and test counts are reported per dataset instance. The analysis is restricted to PII types appearing in the fixed test set. We sort the types by their frequency in the training set and report the ten highest-frequency, ten middle-frequency, and ten lowest-frequency types.

\paragraph{High-frequency types.}
Table~\ref{tab:error-high-frequency} shows the ten highest-frequency PII types. Most high-frequency types achieve strong performance. Personal names reach 0.989, while age, banking business, and lot number reach 0.947, 0.901, and 0.892, respectively. Performance still varies across frequent types, indicating that training frequency alone does not determine how well each PII type is predicted.

\begin{table}[t]
\centering
\scriptsize
\setlength{\tabcolsep}{2.0pt}
\renewcommand{\arraystretch}{0.98}
\begin{tabular*}{\columnwidth}{@{\extracolsep{\fill}}p{0.43\columnwidth}rrrrr@{}}
\toprule
\textbf{PII type}
& \textbf{Train}
& \textbf{Test}
& \textbf{P}
& \textbf{R}
& \textbf{F1} \\
\midrule
Personal names
& 90,028 & 10,126 & 0.983 & 0.995 & 0.989 \\
Internal duties and authorities
& 4,530 & 722 & 0.750 & 0.897 & 0.817 \\
Banking business
& 2,922 & 313 & 0.868 & 0.936 & 0.901 \\
Building and unit number
& 2,877 & 565 & 0.771 & 0.802 & 0.786 \\
Construction execution business
& 2,382 & 268 & 0.685 & 0.874 & 0.768 \\
Central and local government body
& 1,429 & 97 & 0.672 & 0.739 & 0.704 \\
Age
& 1,320 & 189 & 0.965 & 0.929 & 0.947 \\
Housing facility
& 1,181 & 108 & 0.682 & 0.833 & 0.750 \\
Lot number
& 1,173 & 218 & 0.916 & 0.869 & 0.892 \\
Hospital and clinic
& 1,137 & 132 & 0.628 & 0.793 & 0.701 \\
\bottomrule
\end{tabular*}
\caption{Per-type W. Strict Typed precision (P), recall (R), and F1 for high-frequency PII types. Train and test denote the number of annotated instances.}
\label{tab:error-high-frequency}
\end{table}

\paragraph{Middle-frequency types.}
Table~\ref{tab:error-middle-frequency} shows much larger differences among types with similar training frequencies. The \texttt{url} type reaches 0.953 with 161 training instances, whereas fishery reaches only 0.098 with the same number of instances and agriculture receives 0.000 with 158 instances. Trust business also reaches 0.895 with only 143 training instances. These differences indicate that distinctive surface patterns can provide strong cues for prediction even with relatively limited training data, while fine-grained industry and institution types rely more heavily on contextual and semantic information.

\begin{table}[t]
\centering
\scriptsize
\setlength{\tabcolsep}{2.0pt}
\renewcommand{\arraystretch}{0.98}
\begin{tabular*}{\columnwidth}{@{\extracolsep{\fill}}p{0.43\columnwidth}rrrrr@{}}
\toprule
\textbf{PII type}
& \textbf{Train}
& \textbf{Test}
& \textbf{P}
& \textbf{R}
& \textbf{F1} \\
\midrule
Media and press industry
& 166 & 94 & 0.600 & 0.351 & 0.443 \\
Hotel and resort business
& 166 & 5 & 0.094 & 0.200 & 0.128 \\
url
& 161 & 14 & 0.932 & 0.976 & 0.953 \\
Fishery
& 161 & 4 & 0.053 & 0.750 & 0.098 \\
Agriculture
& 158 & 8 & 0.000 & 0.000 & 0.000 \\
Religious organization
& 158 & 68 & 0.524 & 0.902 & 0.663 \\
Financial agency
& 149 & 5 & 0.176 & 0.600 & 0.273 \\
Trust business
& 143 & 83 & 0.854 & 0.940 & 0.895 \\
Credit finance business
& 125 & 31 & 0.722 & 0.419 & 0.531 \\
Cafe
& 125 & 7 & 0.607 & 0.810 & 0.694 \\
\bottomrule
\end{tabular*}
\caption{Per-type W. Strict Typed precision (P), recall (R), and F1 for middle-frequency PII types. Train and test denote the number of annotated instances.}
\label{tab:error-middle-frequency}
\end{table}

\paragraph{Low-frequency types.}
Table~\ref{tab:error-low-frequency} shows the ten lowest-frequency PII types appearing in the test set. Each has only one to three training instances, and all receive 0.000. These cases reflect the long-tailed distribution of the judgment corpus, where some legally relevant PII types occur only rarely. Their limited training support makes reliable prediction difficult.

\begin{table}[t]
\centering
\scriptsize
\setlength{\tabcolsep}{2.0pt}
\renewcommand{\arraystretch}{0.98}
\begin{tabular*}{\columnwidth}{@{\extracolsep{\fill}}p{0.43\columnwidth}rrrrr@{}}
\toprule
\textbf{PII type}
& \textbf{Train}
& \textbf{Test}
& \textbf{P}
& \textbf{R}
& \textbf{F1} \\
\midrule
Stage or pen name
& 3 & 5 & 0.000 & 0.000 & 0.000 \\
Street
& 2 & 2 & 0.000 & 0.000 & 0.000 \\
Password
& 2 & 1 & 0.000 & 0.000 & 0.000 \\
Voting district
& 2 & 1 & 0.000 & 0.000 & 0.000 \\
Exit number
& 2 & 2 & 0.000 & 0.000 & 0.000 \\
Investment products
& 2 & 5 & 0.000 & 0.000 & 0.000 \\
National heritage
& 1 & 6 & 0.000 & 0.000 & 0.000 \\
Corporate registration number
& 1 & 2 & 0.000 & 0.000 & 0.000 \\
Block
& 1 & 1 & 0.000 & 0.000 & 0.000 \\
Vehicle identification number
& 1 & 8 & 0.000 & 0.000 & 0.000 \\
\bottomrule
\end{tabular*}
\caption{Per-type W. Strict Typed precision (P), recall (R), and F1 for low-frequency PII types. Train and test denote the number of annotated instances.}
\label{tab:error-low-frequency}
\end{table}

The results show two recurring patterns. Extremely rare PII types are limited by insufficient training support. Among types with similar training frequencies, types with distinctive surface patterns, such as \texttt{url}, achieve substantially higher performance than fine-grained categories that depend more heavily on context to distinguish them from related types. Training frequency alone therefore does not explain the differences in per-type performance.
\section{Open-weight LLM Evaluation}
\label{app:open-weight-llm}

We evaluate four instruction-tuned open-weight LLMs on the 587-document judgment test set used for the in-domain evaluation. All models are run locally, and no in-context examples are provided. The results reported in Table~\ref{tab:open-weight-llm} use the prompt configuration described below.

\paragraph{Prompting.}
Each document chunk is provided as a separate user message. The system prompt includes the complete fine-grained PII label inventory and instructs the model to preserve the source text exactly while inserting typed PII tags. Table~\ref{tab:open-weight-prompt} shows the prompt used in the experiments, with the full label inventory omitted for compactness. The complete taxonomy is provided in Appendix~\ref{app:PIItaxonomy}.

\begin{table}[t]
\centering
\scriptsize
\setlength{\tabcolsep}{3pt}
\renewcommand{\arraystretch}{1.08}
\begin{tabular}{@{}p{0.20\columnwidth}p{0.73\columnwidth}@{}}
\toprule
\textbf{Role} & \textbf{Prompt} \\
\midrule

System prompt
(original)
&
당신은 한국어 문서의 세부 라벨 비식별화 도구입니다.
아래 라벨 집합에 해당하는 모든 문자열을 정확히
\texttt{<PII type="라벨명">원문 문자열</PII>}
형식으로 감싸세요.
type에는 반드시 아래 집합의 라벨을 철자까지 그대로 사용하세요.
인명·전화번호 같은 통상적 개인정보만 찾지 말고,
종교·시설·조직 내 직책 등 집합에 포함된 모든 종류를 빠짐없이 찾으세요.

판결서에서
\texttt{\#\#\# [라벨링 대상 본문]}
이전의 사건번호·사건명·법원·유형·출처 메타데이터에는
태그를 삽입하지 마세요.
단, 해당 메타데이터와 표식 자체도 생략하지 말고
입력의 첫 글자부터 마지막 글자까지 모두 출력해야 합니다.

원문의 모든 글자, 공백, 줄바꿈, 문장부호와 순서를
추가·삭제·수정하지 마세요.
태그 이름은 항상 \texttt{PII}여야 합니다.
라벨은 오직 \texttt{PII} 태그의 \texttt{type} 속성에만 사용하세요.
라벨 자체를 태그 이름으로 사용하거나
원문을 라벨 placeholder로 치환하지 마세요.
태그 외 설명, Markdown, 코드 펜스를 출력하지 말고
태그가 삽입된 원문 전체만 출력하세요.

\textit{[전체 PII 라벨 집합]}
\\
\midrule

System prompt
(English translation)
&
You are a fine-grained de-identification tool for Korean documents.
Wrap every string corresponding to the label set below exactly in the form
\texttt{<PII type="label">original text</PII>}.
The value of \texttt{type} must exactly match one of the labels in the provided set.
Do not restrict detection to conventional personal information such as names and
phone numbers; identify all categories in the label set, including religion,
facilities, and roles within organizations.

For court judgments, do not insert tags into the case number, case name, court,
type, or source metadata appearing before
\texttt{\#\#\# [라벨링 대상 본문]}.
However, do not omit this metadata or the marker itself; output the complete input
from the first character to the last.

Do not add, delete, or modify any character, whitespace, line break,
punctuation mark, or ordering in the source text.
The tag name must always be \texttt{PII}, and labels must be used only in the
\texttt{type} attribute of the \texttt{PII} tag.
Do not use labels themselves as tag names or replace the source text with label
placeholders.
Output only the full source text with the inserted tags, without explanations,
Markdown, or code fences.

\textit{[Complete PII label inventory]}
\\
\midrule

User message
&
\texttt{\{document chunk\}}
\\

\bottomrule
\end{tabular}
\caption{Prompt configuration used for the open-weight LLM evaluation.}
\label{tab:open-weight-prompt}
\end{table}

\paragraph{Chunking.}
Documents exceeding a model's context limit are divided into chunks that fit within the available context window. Each chunk is submitted as a separate user message under the same system prompt. Including the complete PII inventory in the system prompt reduces the context available for the judgment text and increases the need for chunking on long documents.

\paragraph{Source-preservation validation.}
Because generative models may rewrite, omit, or insert parts of the input, we validate source preservation in addition to measuring de-identification performance. After removing valid predicted PII tags, each generated chunk is compared with the corresponding input text. A document is counted as source-preserved only when every chunk exactly matches the original source text. If any chunk contains an unintended rewrite, deletion, or insertion, the document is counted as not preserved.

Qwen 3.5 2B does not preserve any of the 587 judgments and achieves only 0.004 W. Strict Typed F1. Gemma 4 E2B and Gemma 4 26B-A4B preserve 25.89\% and 27.77\% of the documents, with W. Strict Typed F1 scores of 0.141 and 0.184, respectively. Qwen 3.6 27B performs substantially better, preserving 81.09\% of the documents and reaching 0.483.

To examine whether source modification explains the lower de-identification performance, we additionally evaluate Qwen 3.6 27B on the 476 judgments whose source text is preserved exactly. Its W. Strict Typed F1 increases only to 0.495. This small increase shows that source modification accounts for only a small part of the gap with \sys; typed PII detection remains substantially weaker even when the source text is preserved.

\section{Hardware}
\label{app:hardware}

Model training and evaluation were conducted on three identically configured compute nodes. Each node was equipped with two Intel Xeon Platinum 8468 CPUs, 1.85 TiB of RAM, and eight NVIDIA H100 80GB HBM3 GPUs. Table~\ref{tab:hardware} summarizes the hardware configuration.

\begin{table}[ht]
\centering
\small
\setlength{\tabcolsep}{5pt}
\renewcommand{\arraystretch}{1.08}
\begin{tabular}{@{}ll@{}}
\toprule
Item & Specification \\
\midrule
Number of nodes & 3 \\
CPU & 2$\times$ Intel Xeon Platinum 8468 per node \\
CPU cores & 96 per node \\
RAM & 1.85 TiB per node \\
GPU & 8$\times$ NVIDIA H100 80GB HBM3 per node \\
GPU memory & 81{,}559 MiB per GPU \\
NVIDIA driver & 535.183.06 \\
\bottomrule
\end{tabular}
\caption{Hardware configuration of the compute nodes used for model training and evaluation.}
\label{tab:hardware}
\end{table}
    
\end{document}